\documentclass[12pt]{article}

\usepackage{geometry}

\usepackage{ijcai16}

\usepackage{times}

\usepackage[labelfont=bf]{caption}

\usepackage{amsmath}
\usepackage{amsthm}
\usepackage{amssymb}
\usepackage{float}

\usepackage{enumitem}

\usepackage[dvipsnames]{xcolor}

\usepackage{hyperref}

\hypersetup{
    colorlinks=true,       
    linkcolor=RoyalBlue,          
    citecolor=RoyalBlue,       
    urlcolor=RoyalBlue
}

\usepackage{enumitem}
\setlist[itemize,1]{leftmargin=3em}

\usepackage{tikz}
\usetikzlibrary{arrows}
\usetikzlibrary{arrows.meta}

\usepackage{pgf}
\usetikzlibrary{arrows,automata}
\usetikzlibrary{positioning}

\tikzset{
    state/.style={
           rectangle,
           rounded corners,
           draw=black, very thick,
           minimum height=2em,
           inner sep=2pt,
           text centered,
           },
}

\usepackage{thmtools}
\usepackage{thm-restate}
\usepackage{cleveref}

\declaretheorem[name=Lemma]{lem}

\definecolor{commentRed}{cmyk}{0, 0.90, 0.60, 0}
\definecolor{commentBlue}{cmyk}{0.8, 0, 0.30, 0.1}

\newtheorem{theorem}{Theorem}
\newtheorem{definition}{Definition}

\newtheorem{corollary}{Corollary}

\newtheorem{open question}{Open Question}

\newenvironment{newproof}{\par\vspace*{0.5ex}\noindent\rm {\bf Proof:} }
{
}

\newcommand{\mods}[1]{[\![#1]\!]}
\newcommand{\bel}[1]{[#1]}
\newcommand{\cbel}[1]{[#1]_{\mathrm{c}}}

\newcommand{\AlphaRevR}[1]{$(\alpha{#1}^{\ast}_{\scriptscriptstyle \preceq})$}
\newcommand{\AlphaRevS}[1]{$(\alpha{#1}^{\ast})$}
\newcommand{\BetaRevR}[1]{$(\beta{#1}^{\ast}_{\scriptscriptstyle \preceq})$}
\newcommand{\BetaRevS}[1]{$(\beta{#1}^{\ast})$}
\newcommand{\BetaRevPlusR}[1]{$(\beta{#1 +}^{\ast}_{\scriptscriptstyle \preceq})$}
\newcommand{\BetaRevPlusS}[1]{$(\beta{#1 +}^{\ast})$}
\newcommand{\GammaRevR}[1]{$(\gamma{#1}^{\ast}_{\scriptscriptstyle \preceq})$}

\newcommand{\GammaRevPlusR}[1]{$(\gamma{#1 +}^{\ast}_{\scriptscriptstyle \preceq})$}

\newcommand{\OmegaRevS}[1]{$(\Omega{#1}^{\ast})$}
\newcommand{\BetaCircPlusR}[1]{$(\beta{#1 +}^{\scriptstyle \circ}_{\scriptscriptstyle \preceq})$}

\newcommand{\Leq}[1]{$(\leq{#1})$}

\newcommand{\CRevR}[1]{$(\mathrm{C}{#1}^{\ast}_{\scriptscriptstyle \preceq})$}
\newcommand{\CRevS}[1]{$(\mathrm{C}{#1}^{\ast})$}
\newcommand{\CCircR}[1]{$(\mathrm{C}{#1}^{\scriptstyle \circ}_{\scriptscriptstyle \preceq})$}

\newcommand{\IndRevR}{$(\mathrm{P}^{\ast}_{\scriptscriptstyle \preceq})$}
\newcommand{\IndRevS}{$(\mathrm{P}^{\ast})$}
\newcommand{\IndRevPlusR}{$(\mathrm{P +}^{\ast}_{\scriptscriptstyle \preceq})$}

\newcommand{\IndCircR}{$(\mathrm{P}^{\circ}_{\scriptscriptstyle \preceq})$}

\newcommand{\KRev}[1]{$(\mathrm{K}{#1}^{\ast})$}

\newcommand{\iKRev}[1]{$(\mathrm{iK}{#1}^{\ast})$}
\newcommand{\DR}{$(\mathrm{DR}^{\ast})$}
\newcommand{\DO}{$(\mathrm{DO}^{\ast})$}
\newcommand{\DI}{$(\mathrm{DI}^{\ast})$}
\newcommand{\DF}{$(\mathrm{DF}^{\ast})$}
\newcommand{\iDR}{$(\mathrm{iDR}^{\ast})$}
\newcommand{\iDO}{$(\mathrm{iDO}^{\ast})$}
\newcommand{\iDI}{$(\mathrm{iDI}^{\ast})$}
\newcommand{\iDF}{$(\mathrm{iDF}^{\ast})$}
\newcommand{\IIA}{$(\mathrm{IIA}^{\ast})$}
\newcommand{\iPres}{$(\mathrm{iPres}^{\ast})$}

\newcommand{\WPURevPlus}{$(\mathrm{WPU+}^{\ast})$}
\newcommand{\SPURevPlus}{$(\mathrm{SPU+}^{\ast})$}
\newcommand{\Rec}{$(\mathrm{Rec}^{\ast})$}

\newcommand{\SepRevR}{$(\mathrm{Sep}^{\ast}_{\scriptscriptstyle \preceq})$}
\newcommand{\SepRevS}{$(\mathrm{Sep}^{\ast})$}
\newcommand{\EquivalenceRevS}{$(\mathrm{Eq}^{\ast})$}
\newcommand{\EquivalenceRevR}{$(\mathrm{Eq}^{\ast}_{\scriptscriptstyle \preceq})$}

\newcommand{\EquivalenceCircR}{$(\mathrm{Eq}^{\circ}_{\scriptscriptstyle \preceq})$}

\title{Extending the Harper Identity to Iterated Belief Change}
\author{Richard Booth \\
Cardiff University \\
Cardiff, UK \\
boothr2@cardiff.ac.uk 
\And
Jake Chandler \\
La Trobe University \\
Melbourne, Australia \\
jacob.chandler@latrobe.edu.au}

\begin{document}


 \fontsize{12}{14}\selectfont 
\pagestyle{plain}
\onecolumn

\begin{centering}

{\huge On Strengthening the Logic of Iterated Belief Revision:\\ Proper Ordinal Interval Operators}

\vspace{2em}

\begin{tabular}{ c c c }
  {\Large Richard Booth}  & ~~~~~~~~~~~~~~~~~~~~  & {\Large Jake Chandler} \\
  {\large Cardiff University} & ~ & {\large La Trobe University} \\
  {\large Cardiff, UK} & ~ & {\large Melbourne, Australia} \\
  {\large boothr2@cardiff.ac.uk} & ~ & {\large jacob.chandler@latrobe.edu.au} \\  
\end{tabular}

\vspace{3.5em}

\end{centering}

\begin{quote}

{\large {\bf Abstract} Darwiche and Pearl's seminal 1997 article outlined a number of baseline principles for a logic of iterated belief revision. These principles, the DP postulates, have been supplemented in a number of alternative ways. Most of the suggestions made have resulted in a form of `reductionism' that identifies belief states with orderings of worlds. However this position has  recently been criticised as being unacceptably strong. Other proposals, such as the popular principle (P), aka `Independence', characteristic of `admissible' revision operators, remain commendably more modest. In this paper, we supplement both the DP postulates and (P) with a number of novel conditions. While the DP postulates constrain the relation between a prior and a posterior conditional belief set, our new principles notably govern the relation between two posterior conditional belief sets obtained from a common prior by different revisions. We show that operators from the resulting family, which subsumes both lexicographic and restrained revision, can be represented as relating belief states that are associated with   a `proper ordinal interval' (POI) assignment, a structure more fine-grained than a simple ordering of worlds. We close the paper by noting that these  operators satisfy iterated versions of a large number of AGM era postulates, including Superexpansion, that are not sound for admissible operators in general. }
\end{quote}

~

~


\section{Introduction}


Darwiche \& Pearl's \shortcite{darwiche1997logic} seminal paper put forward a number of now popular baseline principles of iterated belief revision. These principles, the DP postulates, have been strengthened in various manners. Most proposals for doing so--such as natural \cite{boutilier1996iterated}, lexicographic \cite{nayak2003dynamic}, and  restrained \cite{booth2006admissible} revision (see \cite{peppas2014panorama} for an overview)--have yielded sets of principles strong enough to entail the following strong `reductionist' principle:  the set of beliefs held by an agent after a sequence of two revisions is fully determined by the agent's single-step revision dispositions. This thesis can alternatively be cashed out in terms of an identification of belief states, the relata of the revision function, with total preorders (TPO's) over possible worlds. Booth \& Chandler \shortcite{booth2017irreducibility} have however recently provided considerations that suggest this reductionist position to be too strong. Other supplements to the DP postulates, however, have fallen short of having such a consequence. This is true of the popular principle termed `(P)' by Booth \& Meyer \shortcite{booth2006admissible} and `Independence' by Jin \& Thielscher \shortcite{jin2007iterated}, which, together with the DP postulates, characterises the family of `admissible' revision operators that includes both lexicographic and restrained operators but excludes Boutilier's natural ones.

In this paper, we supplement both the DP postulates and (P)~with a number of novel conditions. While the DP postulates constrain the relation between a prior and a posterior conditional belief set, our new principles notably govern the relation between two posterior conditional belief sets obtained from a common prior by different revisions. We take as our foil two postulates of this variety considered by Booth \& Meyer \shortcite{booth2011revise}. These characterised a family of  {\em non-prioritised} revision operators, for which they offered a representation in terms of what we shall call `proper ordinal interval assignments'. Here, we show that these two postulates become  implausible in the context of {\em prioritised} revision, which is the focus of the present paper. First of all, they turn out to characterise lexicographic revision when one supplements the remaining postulates of Booth \& Meyer, i.e.~(P) and the DP postulates, with the AGM postulate of Success. Secondly, they fall prey to an intuitive class of counterexample. After noting this, we then consider two, more plausible, weaker counterparts that have not yet been discussed in the literature. We show that these can be obtained from Booth \& Meyer's construction by adding a `naturalisation' step. This is essentially an application of Boutillier's natural revision operation to the posterior TPO obtained by Booth \& Meyer's method of non-prioritised revision. We call the resulting family of iterated revision operators, which subsumes both lexicographic and restrained revision operators, the family of  `Proper Ordinal Interval (POI)' operators and offer semantic and syntactic characterisations thereof. We close the paper by noting that POI revision operators satisfy iterated versions of a large number of AGM era postulates, including Superexpansion.
\sloppy

The plan of the remainder of the article is as follows. In \hyperref[s:Preliminaries]{`Preliminaries'}, we first introduce some basic terminology and definitions. \hyperref[s:Twoprinciples]{`Two principles of non-prioritised revision'} recapitulates Booth \& Meyer's framework and introduces its two key postulates. These postulates are then critically discussed in \hyperref[s:Theprinciples]{`The principles in a prioritised setting'}. \hyperref[s:Success]{`Success via naturalisation'} outlines our construction of the POI family of operators. In \hyperref[s:Twoweakerprinciples]{`Two weaker principles'}, we discuss the weakenings of Booth \& Meyer's postulates that are satisfied by the members of our new family. In \hyperref[s:characterisation]{`Characterisations of POI operators'}, the family is then characterised semantically 
and syntactically, in two different manners. 
We wrap up the paper with a fairly substantial  discussion, in \hyperref[s:Iteratedversions]{`Iterated versions of AGM era postulates'}, of the extent to which the members of the POI family satisfy extensions of various strong AGM era postulates to the iterated case. We then conclude in \hyperref[s:Conclusions]{`Conclusions and further work'}.
With one or two exceptions, all proofs have been relegated to the \hyperref[s:Appendix]{appendix}.
%

\section{Preliminaries}\label{s:Preliminaries}


The beliefs of an agent are represented by a {\em belief state} $\Psi$. $\Psi$ determines a {\em belief set} $\bel{\Psi}$, a deductively closed set of sentences, drawn from a finitely generated propositional, truth-functional language $L$.  Logical equivalence is denoted by $\equiv$ and the set of logical consequences of $\Gamma\subseteq L$ by $\mathrm{Cn}(\Gamma)$. The set of propositional worlds is denoted by $W$, and the set of models of a given sentence $A$ is denoted by $\mods{A}$. We shall occasionally use $x$ to denote, not the world $x$, but an arbitrary sentence whose set of models is $\{x\}$.

In terms of belief dynamics, our principal focus is on iterated revision--rather than contraction--operators, which return, for any prior belief state $\Psi$ and consistent sentence $A$, the posterior belief state $\Psi \ast A$ that results from an adjustment of $\Psi$ to accommodate the inclusion of $A$ in $\bel{\Psi}$. 

The function $\ast$ is assumed to satisfy the AGM postulates \cite{alchourron1985logic,darwiche1997logic}--henceforth `AGM', for short--which notably include the postulate of Success ($A\in\bel{\Psi\ast A}$). This ensures the following convenient representability of single-shot revision: each $\Psi$ has associated with it a total preorder  $\preceq_\Psi$ over $W$ such that $\mods{\bel{\Psi\ast A}} = \min(\preceq_\Psi, \mods{A})$ \cite{katsuno1991propositional,grove1988two}. This ordering is sometimes interpreted in terms of relative `(im)plausibility', so that $x\preceq_{\Psi} y$ iff $x$ is considered more `plausible' than $y$ in state $\Psi$. In this context, Success corresponds to the requirement that $\min(\preceq_{\Psi \ast A}, W) \subseteq \mods{A}$.

The single-shot revision dispositions associated with $\Psi$ can also be represented by a `{\em conditional belief set}' $\cbel{\Psi}$. This set extends the belief set $\bel{\Psi}$ by further including various `conditional beliefs', of the form $A \Rightarrow B$, where $\Rightarrow$ is a non-truth-functional conditional connective. This is achieved by means of the so-called Ramsey Test, according to which $A \Rightarrow B \in \cbel{\Psi}$ iff $B \in \bel{\Psi \ast A}$.

Following convention, we shall call principles couched in terms of belief sets `syntactic', and principles couched in terms of TPOs `semantic'. The principles that we will discuss will be given in both types of format, with the distinction reflected in the nomenclature by the use of a subscript `$\preceq$' to denote  semantic principles.

We shall also be touching on a broader class of non-prioritised iterated `revision' operators, for which Success does not necessarily hold. These will be denoted by the symbol $\circ$. To avoid ambiguity, we will follow a convention of superscripting every principle governing a belief change operator with the relevant operator symbol (here: $\ast$ or $\circ$).

Finally, $\ast$ will be assumed to satisfy a principle of irrelevance of syntax that we shall call `Equivalence':
\begin{tabbing}
\=BLAHI: \=\kill
\>  \EquivalenceRevS \> If $A\equiv B$ and $C\equiv D$, then $\bel{(\Psi\ast  A)\ast C}=\bel{(\Psi\ast B)\ast D}$\\[-0.25em]
\end{tabbing} 
\vspace{-1em}

\noindent or semantically
\begin{tabbing}
\=BLAHI: \=\kill
\>  \EquivalenceRevR \> If $A\equiv B$, then $\preceq_{\Psi\ast A}=\preceq_{\Psi\ast B}$ 
\\[-0.25em]
\end{tabbing} 
\vspace{-1em}

\noindent as well as the DP postulates, which constrain the belief set resulting from two successive revisions, or, equivalently, the conditional belief set resulting from a single revision:
\begin{tabbing}
\=BLAHI: \=\kill
\> \CRevS{1} \> If $A \in \mbox{Cn}(B)$, then $\bel{(\Psi \ast A) \ast B} = \bel{\Psi \ast B}$  \\[0.1cm]

\> \CRevS{2} \> If $\neg A \in \mbox{Cn}(B)$, then $\bel{(\Psi \ast A) \ast B} = \bel{\Psi \ast B}$ \\[0.1cm]

\> \CRevS{3} \> If $A \in \bel{\Psi \ast B}$, then $A \in \bel{(\Psi \ast A) \ast B}$  \\[0.1cm]

\> \CRevS{4} \> If $\neg A \not\in \bel{\Psi \ast B}$, then $\neg A \not\in \bel{(\Psi \ast A) \ast B}$\\[-0.25em]
\end{tabbing} 
\vspace{-1em}

\noindent whose semantic counterparts are given by:
\begin{tabbing}
\=BLAHI: \=\kill
\> \CRevR{1}  \> If $x,y \in \mods{A}$, then $x \preceq_{\Psi \ast A} y$ iff  $x \preceq_\Psi y$\\[0.1cm]
\> \CRevR{2} \> If $x,y \in \mods{\neg A}$, then $x \preceq_{\Psi \ast A} y$ iff  $x \preceq_\Psi y$\\[0.1cm]
\> \CRevR{3} \> If $x \in \mods{A}$, $y \in \mods{\neg A}$ and $x \prec_\Psi y$, then $x \prec_{\Psi \ast A}y$ \\[0.1cm]
\> \CRevR{4} \> If $x \in \mods{A}$, $y \in \mods{\neg A}$ and $x \preceq_\Psi y$, then $x \preceq_{\Psi \ast A} y$ \\[-0.25em]
\end{tabbing} 
\vspace{-1em}

\noindent In fact we shall further assume that $\ast$ satisfies the principle \IndRevS, which strengthens both \CRevS{3} and \CRevS{4}: 
\begin{tabbing}
\=BLAHI: \=\kill
\> \IndRevS \> If $\neg A \not\in \bel{\Psi \ast B}$, then $A \in \bel{(\Psi \ast A) \ast B}$ \\[-0.25em]
\end{tabbing} 
\vspace{-1em}
Its semantic counterpart is given by:
\begin{tabbing}
\=BLAHI: \=\kill
\>  \IndRevR \> If $x \in \mods{A}$, $y \in \mods{\neg A}$ and $x \preceq_{\Psi} y$, then $x \prec_{\Psi \ast A} y$ \\[-0.25em]
\end{tabbing} 
\vspace{-1em}

\noindent Satisfaction of AGM, \EquivalenceRevS, \CRevS{1}, \CRevS{2} and \IndRevS~means that $\ast$ is an `{\em admissible}' revision operator, in the sense of \cite{booth2006admissible}.

The constraints considered so far are notably satisfied by two well-known kinds of revision operators: restrained operators and lexicographic operators.\footnote{Note the use of the plural here: we speak of restrained/lexicographic operat{\em ors}. It is of course customary, in the literature, to refer to {\em the} restrained/lexicographic operat{\em or}. However, this way of speaking is only appropriate to the extent that belief states are identifiable with TPOs.}  In semantic terms, these both promote the minimal $A$-worlds in the prior TPO to become minimal worlds in the posterior TPO. Regarding the rest of the ordering, restrained revision operators preserve the strict ordering $\prec_{\Psi}$ while additionally making every $A$-world $x$ strictly lower ranked than every $\neg A$-world $y$ for which $x\sim_{\Psi}y$ (where $\sim_{\Psi}$ is the symmetric closure of $\preceq_{\Psi}$), so that $x \preceq_{\Psi \ast A} y$ iff: (i) $x \in \min(\preceq_{\Psi}, \mods{A})$, or (ii) $x, y \notin \min(\preceq_{\Psi}, \mods{A})$ and either (a) $x \prec_{\Psi} y$ or (b) $x \sim_{\Psi} y$ and ($x\in\mods{A}$ or $y\in\mods{\neg A}$). 
%
%
Lexicographic revision operators make every $A$-world lower ranked than every $\neg A$-world, while preserving the ordering within each of $\mods{A}$ and $\mods{\neg A}$, so that $x \preceq_{\Psi \ast A} y$ iff:  (i) $x\in\mods{A}$ and $y\in\mods{\neg A}$, or  (ii) ($x\in\mods{A}$ iff $y\in\mods{A}$) and $x \preceq_{\Psi} y$. 
%
Natural revision operators, however, fail to satisfy \IndRevS ~and are thus not members of the family of admissible revision operators. These operators simply promote the minimal $A$-worlds to be $\preceq_{\Psi \ast A}$-minimal, while leaving everything else unchanged, so that $x \preceq_{\Psi \ast A} y$ iff: (i) $x \in \min(\preceq_{\Psi}, \mods{A})$, or 
(ii) $x, y \notin \min(\preceq_{\Psi}, \mods{A})$ and $x \preceq_{\Psi} y$. 
%
%


\section{Two principles of non-prioritised revision}\label{s:Twoprinciples}


The DP postulates, as well as \IndRevS, constrain the relation between a prior conditional belief set on the one hand, and a posterior one on the other. But one might wonder what kinds of constraints govern the relation between {\em two posterior conditional belief sets} obtained from a common prior by different revisions. 

To the best of our knowledge, the only two articles to consider principles of this nature are  \cite{booth2011revise} and, more briefly,  \cite{schlechta1996distance}. In the former, a slightly more general form of the following pair of syntactic principles is discussed:

\begin{tabbing}
\=BLAHI: \=\kill
\> \BetaRevPlusS{1}  \> If $A \not\in \bel{(\Psi \ast A) \ast B}$, then $A \not\in \bel{(\Psi \ast C) \ast B}$ \\[0.1cm]
\> \BetaRevPlusS{2} \> If $\neg A \in \bel{(\Psi \ast A) \ast B}$, then $\neg A \in \bel{(\Psi \ast C) \ast B}$ \\[-0.25em]
\end{tabbing} 
\vspace{-1em}

\noindent whose semantic counterparts are given by:
\begin{tabbing}
\=BLAHI: \=\kill
\>\BetaRevPlusR{1}  \> If $x \in \mods{A}$, $y \in \mods{\neg A}$ and $y \preceq_{\Psi \ast A} x$, then $y \preceq_{\Psi \ast C} x$ \\[0.1cm]
\> \BetaRevPlusR{2} \> If $x \in \mods{A}$, $y \in \mods{\neg A}$ and $y \prec_{\Psi \ast A} x$, then   $y \prec_{\Psi \ast C} x$ \\[-0.25em]
\end{tabbing} 
\vspace{-1em}

\noindent On the relative plausibility interpretation of $\preceq$, the latter can be informally glossed as follows: if (i) there exists some potential evidence, consistent with a world $x$ but not with a world $y$, such that $x$ would be considered no more plausible than (respectively: strictly less plausible than) $y$ after receiving it,  then (ii) there is no potential evidence {\em whatsoever} that would lead $x$ to be considered more plausible than (respectively: at least as plausible as)  $y$.

It is easy to see that, on the assumption that $\preceq_{\Psi \ast \top} = \preceq_{\Psi}$ (which follows from \CRevR{1}), these respectively generalise \CRevR{3} and \CRevR{4}, which correspond to the special cases in which $C$ is a tautology.

These postulates can be interpreted in a number of ways. One way is in terms of the  binary relations (over consistent sentences in $L$) of {\em overrules} and {\em strictly overrules} \cite{booth2011revise}. We say $B$ overrules $A$ (in $\Psi$) iff $A \not\in \bel{(\Psi \ast A) \ast B}$, while $B$ strictly overrules $A$ (in $\Psi$) iff $\neg A \in \bel{(\Psi \ast A) \ast B}$.\footnote{Incidentally, the first relation also corresponds to the condition under which Chandler \shortcite{Chandler2017-CHAPCA-11} proposed that one takes $B$ to provide a reason to not believe $A$. The second relation is related to the condition under which he claimed one takes $B$ to provide a reason to believe $\neg A$ \cite{Chandler2013-CHATFA-3}.} Then \BetaRevPlusS{1} says that, if $B$ overrules $A$ in $\Psi$, then $A$ will not be believed following any sequence of two revisions starting in $\Psi$ ending with $B$, while \BetaRevPlusS{2} says that, if $B$ strictly overrules $A$ in $\Psi$, then $A$ will be rejected following any  such sequence of two revisions. 

We noted above that it was a more {\em general} form of  \BetaRevPlusS{1},  \BetaRevPlusS{2} and their semantic counterparts that interested Booth \& Meyer. The reason for this is that their topic of interest was not in fact $\ast$, but rather a more {\em general} kind of operator: a  {\em non-prioritised} ``revision'' operator $\circ$, which does not necessarily satisfy the Success postulate. They showed that these operators could be represented as relating belief states to which a certain type of structure is associated. We provide in what follows a brief overview of their framework. First, some key definitions:

\begin{definition}\label{def:POI}
$\leq$ is a {\em proper ordinal interval (POI) assignment} to $W$ iff it is a relation over $W^{\pm} = \{w^{i} \mid w \in W \mbox{ and } i \in \{-, +\}\}$ such that: 
\begin{tabbing}
\=BLAHI: \=\kill
\> \Leq{1} \>  $\leq$ is a TPO \\[0.1cm]

\> \Leq{2} \> $x^{+} < x^{-}$\\[0.1cm]

\> \Leq{3} \> $x^{+} \leq y^{+}$ iff $x^{-} \leq y^{-}$. \\[-0.25em]
\end{tabbing} 
\vspace{-1em}
\end{definition}

\begin{definition}\label{def:faithful}
Where $\preceq$ is a TPO over $W$ and $\leq$ is a POI assignment to $W$, we say that $\leq$ is {\em faithful} to $\preceq$ iff it satisfies:
\begin{tabbing}
\=BLAHI: \=\kill
\> \Leq{4} \>  $x^{+} \leq y^{+}$ iff $x \preceq y$. \\[-0.25em]
\end{tabbing} 
\vspace{-1em}
\end{definition}

\noindent Booth \& Meyer then assumed that each belief state $\Psi$ is associated, not only with a TPO $\preceq_\Psi$, but with a POI assignment $\leq_{\Psi}$ that is faithful to it (they remained agnostic as to whether states are to be identified with POI assignments; we will follow suit). 
This assignment  was then taken to determine the agent's posterior TPO upon revision by $A$, i.e.~$\preceq_{\Psi \circ A}$, in the following manner: 

\begin{definition}\label{def:nonprioritisePOI}
$\circ$ is a {\em non-prioritised POI revision operator} iff $\circ$ is a function from state-sentence pairs to states, such that for every state $\Psi$ there is a POI assignment $\leq_{\Psi}$ such that, for any sentence $A$, $x \preceq_{\Psi \circ A} y$ iff $r_A(x) \leq_\Psi r_A(y)$, where 
\[
r_A(x) =
\left\{
\begin{array}{ll}
x^+ & \textrm{if } x \in \mods{A} \\
x^- & \textrm{if } x \in \mods{\neg A}. 
\end{array}
\right.
\]  
\end{definition}

\noindent General forms of our principles  \BetaRevPlusR{1} and \BetaRevPlusR{2} turn out to play a key role in this model. Indeed, Booth \& Meyer \shortcite[Theorem 1]{booth2011revise} show that $\circ$ is a non-prioritised POI revision operator if and only if it satisfies \CCircR{1}, \CCircR{2}, \IndCircR,  \BetaCircPlusR{1} and \BetaCircPlusR{2}, where these principles are obtained from their counterparts for (prioritised) revision in the obvious manner, by substituting the $\circ$ symbol for $\ast$.

Non-prioritised POI revision operators can helpfully be understood diagrammatically. Figure \ref{fig:POI} represents a proper ordinal interval assignment that is faithful to $x \prec_{\Psi} y \prec_{\Psi} z$. The left and right interval endpoints respectively represent the positive $(\cdot)^+$ and negative $(\cdot)^-$ counterparts of each world. Figure \ref{fig:POINPR} represents, by means of the filled circles, the TPO resulting from the corresponding  non-prioritised revision by $y\vee z$, i.e., $x \sim_{\Psi \circ  y \vee z} y \prec_{\Psi \circ y \vee z} z$. It also illustrates failure of Success, since $x\in\min(\preceq_{\Psi\circ y\vee z}, W)$.

\begin{figure}[H]
  \begin{minipage}[b]{0.45\linewidth}
\centering
\begin{tikzpicture}[shorten >=-3pt,shorten <=-3pt]
\scalebox{1}{
   \draw (.5,1.5) node {$x$};
   \draw[o-o] (1.5,1.5) -- (2.5,1.5);
   \draw (.5,1) node {$y$};
   \draw[o-o] (2.55,1) -- (3.5,1);
    \draw (.5,.5) node {$z$};
    \draw[o-o] (3,.5) -- (4,.5);
}
\end{tikzpicture}
\caption{proper ordinal interval assignment} 
\label{fig:POI}
\end{minipage}
\hspace{0.05\linewidth}
\begin{minipage}[b]{0.45\linewidth}
\centering
\begin{tikzpicture}[shorten >=-3pt,shorten <=-3pt]
\scalebox{1}{
   \draw (.5,1.5) node {$x$};
   \draw[o-*] (1.5,1.5) -- (2.5,1.5);
   \draw (.5,1) node {$y$};
   \draw[*-o] (2.55,1) -- (3.5,1);
    \draw (.5,.5) node {$z$};
    \draw[*-o] (3,.5) -- (4,.5);
}
\end{tikzpicture}
\caption{posterior TPO  after non-prioritised POI revision by $y \vee z$ } 
\label{fig:POINPR}
\end{minipage}
\end{figure}

\noindent We note that lexicographic revision operators are special cases of this family in which $x^+ <_\Psi y^-$ for all $x,y \in W$. 


\section{The principles in a prioritised setting}\label{s:Theprinciples}


In spite of their arguable appropriateness in a non-prioritised setting, \BetaRevPlusR{1} and \BetaRevPlusR{2} prove to be problematically strong when one imposes Success. 

For one, it turns out that, in such a context the only kind of operators satisfying \BetaRevPlusR{1} are lexicographic revision operators, and hence that  \BetaRevPlusR{1} imposes the reductionist assumption that we have suggested is objectionable. Indeed:

\begin{theorem}
\label{beta1lex}
Let $\ast$ be a revision operator satisfying AGM and \BetaRevPlusS{1}. Then it also satisfies the Recalcitrance property \cite{nayak2003dynamic}:
\begin{tabbing}
\=BLAHI: \=\kill
\>  \Rec \>  If $A \wedge B$ is consistent, then $A \in \bel{(\Psi \ast A) \ast B}$. \\[-0.25em]
\end{tabbing} 
\vspace{-1em}
\end{theorem}

\begin{newproof}
If $A \wedge B$ is consistent, then $A \in \bel{(\Psi \ast A \wedge B) \ast B}$ from AGM. Then $A \in \bel{(\Psi \ast A) \ast B}$ by \BetaRevPlusS{1}.\footnote{If one assumes consistency of revision inputs, it is trivial to show that the implication also runs the other way, so that \Rec~and \BetaRevPlusS{1} are then equivalent, given AGM: Suppose $A\in \bel{(\Psi\ast C)\ast B}$. Since we thereby implicitly assume consistency of $B$, $A\wedge B$ must also be consistent (as, by AGM, $A\wedge B\in \bel{(\Psi\ast C)\ast B}$ and $\bel{(\Psi\ast C)\ast B}$ is consistent). Hence, by \Rec, $A \in \bel{(\Psi \ast A) \ast B}$.} \qed
\end{newproof}

\vspace{0.5em}

\noindent Since we know (see, e.g., \cite{booth2011revise,nayak2003dynamic}),  that  lexicographic revision operators are the only admissible operators satisfying \Rec, we obtain the following corollary, which also gives us an alternative characterisation of lexicographic revision operators:

\vspace{-0em}

\begin{corollary}\label{cor:LexChar}
The only operators satisfying AGM, \CRevS{1}, \CRevS{2}  and \BetaRevPlusS{1} are lexicographic revision operators.\footnote{What about \BetaRevPlusR{2}? Lexicographic revision satisfies it trivially, since it satisfies: If $x \in \mods{A}$, $y \in \mods{\neg A}$, then $x \prec_{\Psi \ast A} y$. We can analogously show that it implies, given AGM, the following weakening of \Rec: If $A \wedge B$ is consistent, then $\neg A \notin \bel{(\Psi \ast A) \ast B}$. But this is too weak to allow us to recover \Rec~and indeed, \BetaRevPlusR{2} is not uniquely satisfied by lexicographic revision.
}
\end{corollary}

\noindent These principles also face a class of direct counterexamples that match  the following general pattern: $A$ provides a defeasible reason to believe $\neg B$ (for example, let $B$ = `She missed the target at 5 yards' and A = `She is a pro archer') and $C$ is equivalent to the conjunction of $A$ and a defeater for $A$'s support for $\neg B$ (for example, let $C$ = `She is a pro archer but isn't wearing her glasses'). Under these conditions, it can plausibly be the case that $\neg A \in \bel{(\Psi \ast A) \ast B}$ but $A \in \bel{(\Psi \ast C) \ast B}$, contradicting both principles. 

This negative result raises the following question: Is there any way to weaken \BetaRevPlusR{1} and \BetaRevPlusR{2} to allow a wider, but intuitively plausible family of iterated prioritised revision operators? The answer, as we will now show, is `yes'.


\section{Success via naturalisation}\label{s:Success}


The guiding idea in what follows is to take the family of operators discussed in the section before last and ensure satisfaction of Success, not by adding the principle to the list of characteristic postulates but rather by minimally transforming the TPO associated with the posterior belief state by means of an operation analogous to natural revision.

More precisely, the proposal is to define $\ast$ as the {\em composition} of a non-prioritised POI revision operator $\circ$ and a natural revision operator ${\scriptstyle \,\boxplus\,}$:
\[
\preceq_{\Psi \ast A} = \preceq_{(\Psi \circ A) {\scriptscriptstyle \,\boxplus\,} A} 
\]
\noindent Recalling the definition of natural revision in \hyperref[s:Preliminaries]{`Preliminaries'}, we can equivalently say:

\begin{definition}\label{def:naturalisation}

$\ast$ is a {\em naturalisation} of $\circ$ iff: 
\begin{itemize}
\item[] $x \preceq_{\Psi \ast A} y$ iff either 
\begin{itemize}
\item[] (i) $x \in \min(\preceq_{\Psi\circ A}, \mods{A})$, or 

\item[] (ii) $x, y \notin \min(\preceq_{\Psi\circ A}, \mods{A})$ and $x \preceq_{\Psi\circ A} y$. 
\end{itemize} 
\end{itemize} 
We use $\mathbb{N}(\ast, \circ)$ to denote the fact that this relation obtains between the two functions.
\end{definition}


\begin{definition}\label{def:PoiOperator}
$\ast$ is a {\em proper interval order (POI)} revision operator iff  $\mathbb{N}(\ast, \circ)$ for some non-prioritised POI revision operator $\circ$.
\end{definition}

\noindent This kind of suggestion generalises one that was made in \cite{booth2006admissible}, in which restrained revision operators were shown to be naturalisations of a particular class of non-prioritised revision operators due to Papini \shortcite{Papini2001}. Indeed, the latter satisfy: $x\preceq_{\Psi \circ A} y$  iff (a) $x \prec_{\Psi} y$ or (b) $x \sim_{\Psi} y$ and [$x\in\mods{A}$ or $y\in\mods{\neg A}$]. These conditions, of course, simply correspond to (ii)(a) and (ii)(b) in the definition of restrained revision operators given in \hyperref[s:Preliminaries]{`Preliminaries'}. The proposal is also somewhat reminiscent of the manner in which the Levi Identity \cite{levi1977subjunctives} treats non-iterated revision as the composition of a contraction and an expansion
 ($\bel{\Psi \ast A} = \bel{\Psi\div \neg A} + A$), with our natural revision step ${\scriptstyle \,\boxplus\,}$ playing the role of the expansion step $+$ .


\begin{figure}[H]
\begin{center}
\begin{tikzpicture}[x=0.75pt,y=0.75pt,yscale=1,xscale=1]
\scalebox{1}{

\draw (5,5) node  [align=left] (a1) [scale=1.25] {states};
\draw (5,55) node  [align=left] (a2) [scale=1.25] {POIs};
\draw (5,105) node  [align=left] (a3) [scale=1.25] {TPOs};
\draw (5,160) node  [align=left] (a4) [scale=1.25] {sets};

\draw (105,5)  node  (b1) [scale=1.25] {$\Psi$};
\draw (105,55) node  (b2)  [scale=1.25] {$\leq _{\Psi }$};
\draw (105,105) node  (b3) [scale=1.25] {$\preceq _{\Psi }$};
\draw (105,160) node (b4)  [scale=1.25] {$[ \Psi ]$};

\draw (205,5) node(c1) [scale=1.25]  {$\Psi \circ A$};
\draw (205,55)  node (c2)  [scale=1.25] {$\leq _{\Psi \circ A}$};
\draw (205,105) node (c3) [scale=1.25] {$\preceq _{\Psi \circ A}$};
\draw (205,160) node (c4)  [scale=1.25] {$[ \Psi \circ A]$};

\draw (305,5)  node (d1)  [scale=1.25] {$\Psi \ast A$};
\draw (305,55)  node  (d2) [scale=1.25] {$\leq _{\Psi \ast A}$};
\draw (305,105) node  (d3) [scale=1.25]  {$\preceq _{\Psi \ast A}$};
\draw (305,160) node (d4) [scale=1.25]  {$[ \Psi \ast A]$};

\path[shorten >=0.25cm,shorten <=0.25cm,->] (b1) edge (b2);
\path[shorten >=0.25cm,shorten <=0.25cm,->] (b2) edge (b3);
\path[shorten >=0.25cm,shorten <=0.25cm,->]  (b3) edge (b4);

\path[shorten >=0.25cm,shorten <=0.25cm,->] (c1) edge (c2);
\path[shorten >=0.25cm,shorten <=0.25cm,->] (c2) edge (c3);
\path[shorten >=0.25cm,shorten <=0.25cm,->]  (c3) edge (c4);

\path[shorten >=0.25cm,shorten <=0.25cm,->] (d1) edge (d2);
\path[shorten >=0.25cm,shorten <=0.25cm,->] (d2) edge (d3);
\path[shorten >=0.25cm,shorten <=0.25cm,->]  (d3) edge (d4);
 
\path[->]  (b1) edge[bend right] node [below]  {$\circ A$} (c1);
\path[->]  (c1) edge[bend right] node [below]  {${\scriptscriptstyle \,\boxplus\,} A$} (d1);
\path[->]  (b1) edge[bend right=60] node [below]  {$\ast A$} (d1);

\path[->]  (b2) edge[bend left, dashed] (c3);

\path[->]  (c3) edge[dashed] (d3);

}
\end{tikzpicture}
\end{center}
\caption{functional dependencies in POI revision}
\label{fig:setup}
\end{figure}
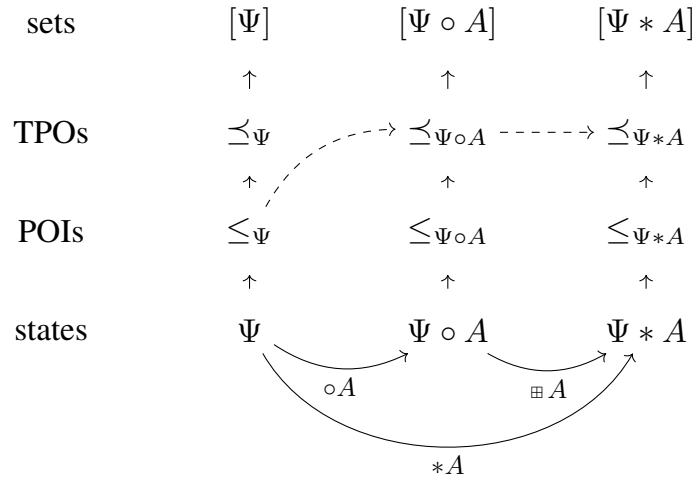

\noindent Figure \ref{fig:setup} provides a general overview of the model, with the various arrows denoting functional determination. From bottom to top, each belief state $\Psi$ is mapped onto a POI $\leq_{\Psi}$. This POI determines a TPO $\preceq_{\Psi}$, such that $x \preceq_{\Psi} y$ iff $x^{+} \leq_{\Psi} y^{+}$. Finally the TPO in turn determines a belief set $\bel{\Psi}$, such that $A\in \bel{\Psi}$ iff $\min(\preceq,W)\subseteq \mods{A}$. These mappings are potentially many-to-one, so that we obtain increasingly coarse descriptions of an agent's beliefs as one moves upwards. From left to right, the function $\circ$ maps the prior belief state $\Psi$ onto an `intermediate' state  $\Psi \circ A$, before the function ${\scriptstyle \,\boxplus\,}$ maps the latter onto the posterior state $\Psi\ast A = (\Psi \circ A) {\scriptstyle \,\boxplus\,} A$. 

We have used dashed arrows to denote some further functional dependencies. The constraints of \cite{booth2011revise} ensure that the prior POI assignment $\leq_\Psi$ determines the `intermediate' TPO $\preceq_{\Psi \circ A}$. Finally, the constraints operating on the function ${\scriptstyle \,\boxplus\,}$ ensure that {\em this} in turn determines the posterior TPO $\preceq_{\Psi\ast A}$. 

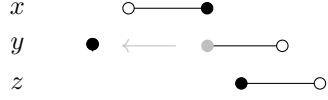
\begin{figure}[H]
\centering
\begin{tikzpicture}[shorten >=-3pt,shorten <=-3pt]
\scalebox{1}{
   \draw (.5,1.5) node {$x$};
   \draw[o-*] (2,1.5) -- (3,1.5);
   \draw (.5,1) node {$y$};
   \draw[*-*] (1.5,1) -- (1.5,1);
   \draw[lightgray, <-] (2,1) -- (2.5,1);
   \draw[arrows={*[lightgray]-o}] (3.05,1) -- (4,1);
    \draw (.5,.5) node {$z$};
    \draw[*-o] (3.5,.5) -- (4.5,.5);
}
\end{tikzpicture}
\caption{posterior TPO after naturalisation}  
\label{fig:POIR}
\end{figure}

This last step is achieved by moving the $\preceq_{\Psi \circ A}$-minimal $A$-worlds to the front of the ordering. Figure \ref{fig:POIR} represents the result of naturalising the posterior TPO depicted in Figure \ref{fig:POINPR}, with $y$ being moved into the leftmost position.

\noindent The naturalisation step ensures that we have $B\in\bel{\Psi\ast A}$ iff $\min(\preceq_{\Psi}, \mods{A})\subseteq\mods{B}$, so AGM will now clearly be satisfied, including Success. Furthermore, the following general fact about naturalisation establishes that the set of POI revision operators is a subset of the set of admissible operators:

\begin{restatable}{prop}{CircPropertyPreservation}
\label{CircPropertyPreservation}
For any iterated revision operators $\circ$ and $\ast$, such that $\mathbb{N}(\ast, \circ)$, if $\circ$ satisfies  \EquivalenceCircR, \CCircR{1}, \CCircR{2} and \IndCircR, then $\ast$ will satisfy \EquivalenceRevR, \CRevR{1}, \CRevR{2} and \IndRevR.
\end{restatable}

\noindent Indeed, we have already noted that non-prioritised POI revision operators satisfy  \CCircR{1}, \CCircR{2} and \IndCircR. Furthermore, Booth \& Meyer show that they also satisfy \EquivalenceCircR.

The family of POI revision operators includes some familiar figures:

\begin{restatable}{prop}{lexresPOI}
\label{lexresPOI}
Both lexicographic and restrained revision operators are POI revision operators.
\end{restatable}

 \noindent Indeed, we have pointed out, at the end of the section titled \hyperref[s:Twoprinciples]{`Two principles of non-prioritised revision'}, that lexicographic revision operators are themselves non-prioritised POI revision operators. Furthermore, since they satisfy Success, they will be identical with their own naturalisations. Regarding restrained revision operators, the result was established in Proposition 14 of \cite{booth2006admissible}: they are, as we noted above, naturalisations of Papini's `reverse' lexicographic revision operators, which are non-prioritised POI revision operators.


\section{Two weaker principles} \label{s:Twoweakerprinciples}


It is easy to see that neither \BetaRevPlusR{1}, nor \BetaRevPlusR{2} are generally satisfied by POI revision operators. Indeed, let $W=\{x, y, z\}$ and $\leq_{\Psi}$ be given as follows: $z^{+} < y^{+} < z^{-} < y^{-}< x^{+} < x^{-} $. Then $y \prec_{\Psi\ast x\vee z} x$, but $x \prec_{\Psi\ast x} y$. However, as we shall see from Proposition \ref{SoundnessAlphas} in the next section, we {\em do} nevertheless obtain the following weakened versions of these principles, which incorporate into their antecedents the further requirement that $x \not\in \min(\preceq, \mods{C})$:

\begin{tabbing}
\=BLAHI: \=\kill
\> \BetaRevR{1} \>  If $x \not\in \min(\preceq, \mods{C})$, $x \in \mods{A}$, $y\in \mods{\neg A}$, and    $y \preceq_{\Psi \ast A} x$, then $y \preceq_{\Psi\ast C} x$ \\[0.1cm]

\> \BetaRevR{2}  \> If $x \not\in \min(\preceq, \mods{C})$, $x \in \mods{A}$, $y\in \mods{\neg A}$, and $y \prec_{\Psi \ast A} x$, then $y \prec_{\Psi\ast C} x$ \\[-0.25em]
\end{tabbing} 
\vspace{-1em}
\noindent Regarding the syntactic counterparts of these principles:
\begin{restatable}{prop}{BetaOneTwoSyntactic}
\label{BetaOneTwoSyntactic}
{\bf (a)} Given AGM, \BetaRevR{1} is equivalent to:
\begin{tabbing}
\=BLAHI: \=\kill
\> \BetaRevS{1}  \>  If $A \not\in \bel{(\Psi \ast A) \ast B}$ and $B \rightarrow \neg A \in \bel{\Psi \ast C}$,    then $A \not\in \bel{(\Psi \ast C) \ast B}$ \\[-0.25em]
\end{tabbing} 
\vspace{-1em}
{\bf (b)} Given AGM, \BetaRevR{2}  is equivalent to:
\begin{tabbing}
\=BLAHI: \=\kill
\> \BetaRevS{2} \> If $\neg A \in \bel{(\Psi \ast A) \ast B}$ and $B \rightarrow \neg A \in \bel{\Psi \ast C}$,   then $\neg A \in \bel{(\Psi \ast C) \ast B}$ \\[-0.25em]
\end{tabbing} 
\vspace{-1em}

\end{restatable}

\noindent These principles are particularly interesting insofar as they avoid the kind of counterexample to \BetaRevPlusR{1} and \BetaRevPlusR{2} that we raised earlier. Indeed, in the scenarios in question, we also intuitively have 
$B \rightarrow \neg A \in \bel{\Psi \ast C}$, rendering them perfectly  consistent with the weaker \BetaRevS{1} and \BetaRevS{2}.

It will turn out to be useful, in the final sections of the paper, to have noted the following equivalent formulations of  \BetaRevR{1} and \BetaRevR{2}:

\begin{restatable}{prop}{BetaGammaEquivalence}
\label{BetaGammaEquivalence}
{\bf (a)} Given \CRevR{2}  and \CRevR{4}, \BetaRevR{1} is equivalent to the conjunction of the following two principles:
\begin{tabbing}
\=BLAHI: \=\kill
\>  \GammaRevR{1} \> If $x \in \mods{A}$, $y \in \mods{\neg A}$ and $y  \preceq_{\Psi \ast A} x$, then   $y  \preceq_{\Psi \ast A\vee C } x$ \\[0.1cm]
\>  \GammaRevR{3} \> If $x \notin \min( \preceq, \mods{C})$, $x \in \mods{A \vee C}$,   $y \in \mods{\neg(A \vee C)}$, and $y  \preceq_{\Psi \ast A \vee C} x$,   then \\
\>\> $y  \preceq_{\Psi \ast C} x$.\\[-0.25em]
\end{tabbing} 
\vspace{-1em}
{\bf (b)} Given \CRevR{1}  and \CRevR{3}, \BetaRevR{2} is equivalent to the conjunction of the following two principles:
\begin{tabbing}
\=BLAHI: \=\kill
\>  \GammaRevR{2} \> If $x \in \mods{A}$, $y \in \mods{\neg A}$ and $y  \prec_{\Psi \ast A} x$, then  $y  \prec_{\Psi \ast A\vee C } x$ \\[0.1cm]
\>  \GammaRevR{4} \> If $x \notin \min( \preceq, \mods{C})$, $x \in \mods{A \vee C}$,     $y \in \mods{\neg(A \vee C)}$ and $y  \prec_{\Psi \ast A \vee C} x$,  then \\
\>\>$y  \prec_{\Psi \ast C} x$. \\[-0.25em]
\end{tabbing} 
\vspace{-1em}
\end{restatable}

\noindent Note that, given the assumption that $\preceq_{\Psi\ast \top} = \preceq_{\Psi}$, which follows from \CRevR{1}, \GammaRevR{1} and \GammaRevR{2} respectively entail \CRevR{3} and \CRevR{4} (let $C=\neg A$). However, none of these four new principles, and hence neither of \BetaRevR{1} and \BetaRevR{2}, are generally sound for admissible operators: 

\begin{restatable}{prop}{wSriTwoSemanticComment}
\label{prop:wSriTwoSemanticComment}
None of \GammaRevR{1}~to \GammaRevR{4}~follows from AGM ,  \CRevR{1},  \CRevR{2}~and \IndRevR~alone.
\end{restatable} 


\section{Characterisations of POI operators}\label{s:characterisation}


We have now identified a number of sound principles for the class of POI revision operators, which, we would like to remind the reader, subsumes both restrained and lexicographic operators. Next, we would like to {\em characterise} it. 


\subsection{Semantic characterisation}\label{ss:Semanticcharacterisation}


For our semantic characterisation, we need to introduce three more postulates, the first two of which are respective strengthenings of \BetaRevR{1}~and \BetaRevR{2}, which can be recovered by setting $z = y$:
\begin{tabbing}
\=BLAHI: \=\kill

\> \AlphaRevR{1} \> If $x \not\in \min(\preceq, \mods{C})$, $x \in \mods{A}$, $y\in \mods{\neg A}$,   $z \preceq_{\Psi}y$ and $y \preceq_{\Psi \ast A} x$, then \\
\>\> $z \preceq_{\Psi \ast C} x$ \\[0.1cm]

\>  \AlphaRevR{2} \> If $x \not\in \min(\preceq, \mods{C})$, $x \in \mods{A}$, $y\in \mods{\neg A}$,   $z \preceq_{\Psi} y$ and $y \prec_{\Psi \ast A} x$, then  \\
\>\> $z \prec_{\Psi \ast C} x$ \\[0.1cm]

\>  \AlphaRevR{3} \>  If $x \not\in \min(\preceq, \mods{C})$, $x \in \mods{A}$, $y\in \mods{\neg A}$,    $z \prec_{\Psi} y$ and $y \preceq_{\Psi \ast A} x$, then  \\
\>\> $z \prec_{\Psi \ast C} x$\\[-0.25em]	

\end{tabbing} 
\vspace{-1em}

\begin{restatable}{prop}{SoundnessAlphas}
\label{SoundnessAlphas}
\AlphaRevR{1}, \AlphaRevR{2} and \AlphaRevR{3} are satisfied by POI revision operators.
\end{restatable}

\noindent These principles can perhaps be viewed as qualified pseudo-`transitivity' principles, if one ignores the subscripts. \AlphaRevR{1} and \AlphaRevR{2} amount to the conjunctions of \BetaRevR{1} and \BetaRevR{2} with the following semantic postulates, respectively:
\begin{tabbing}
\=BLAHI: \=\kill
\>\BetaRevR{3} \> If $z \neq y$, $x \notin \min(\preceq, \mods{C})$, $x \in \mods{A}$, $y \in \mods{\neg A}$,  $z \preceq_{\Psi} y$,  and $y \preceq_{\Psi\ast A} x$,  \\
\>\> then  $z \preceq_{\Psi\ast C} x$\\[0.1cm]

\>\BetaRevR{4}  \> If $z \neq y$, $x \notin \min(\preceq, \mods{C})$, $x \in \mods{A}$, $y \in \mods{\neg A}$, $z \preceq_{\Psi} y$,  and $y \prec_{\Psi\ast A} x$,  \\
\>\> then  $z \prec_{\Psi\ast C} x$\\[-0.25em]
\end{tabbing} 
\vspace{-1em}

\noindent So Proposition \ref{SoundnessAlphas} shows that \BetaRevR{1} and \BetaRevR{2}--and hence, in view of Proposition \ref{BetaGammaEquivalence}, \GammaRevR{1} to \GammaRevR{4}--are sound for POI revision operators. We can now present our main result, which is a semantic characterisation of the family:

\begin{restatable}{theorem}{POIrepresentation}
\label{POIrepresentation}
$\ast$ is a POI revision operator iff it satisfies AGM, \EquivalenceRevR, \CRevR{1}, \CRevR{2}, \IndRevR, \AlphaRevR{1}, \AlphaRevR{2}, and \AlphaRevR{3}.
\end{restatable}

\noindent In conjunction with the results of Booth \& Meyer regarding non-prioritised POI revision operators, Propositions \ref{CircPropertyPreservation} and  \ref{SoundnessAlphas} establish the left-to-right direction of the above claim. For the other direction we need to show that, if $\ast$ satisfies the relevant semantic properties, then there exists a non-prioritised POI revision operator $\circ$ such that $\mathbb{N}(\ast, \circ)$. The construction works as follows: From $\ast$, define $\circ$ by setting, for all $x,y \in W$, $x \preceq_{\Psi \circ A} y$ iff $x \preceq_{\Psi \ast A \vee \neg(x\vee y)} y$. Given \EquivalenceRevR, \CRevR{1} and  \CRevR{2}  this is equivalent to:
\[
x \preceq_{\Psi \circ A} y\
\textrm{iff }
\left\{
\begin{array}{ll}
x \preceq_{\Psi} y & \textrm{if } x \sim^A y \\
x \preceq_{\Psi \ast \neg y} y & \textrm{if } x \triangleleft^A y \\
x \preceq_{\Psi \ast \neg x} y & \textrm{if } y \triangleleft^A x\end{array}
\right.
\]

\noindent where (i) $x\trianglelefteq^A y$ iff $x\in\mods{A}$ or $y\in\mods{\neg A}$, (ii) $x\sim^A y$ iff $x\trianglelefteq^A y$ and $y\trianglelefteq^A x$,  (iii) $x\triangleleft^A y$ iff $x\trianglelefteq^A y$ but not $y\trianglelefteq^A x$.


\subsection{Two syntactic characterisations}
\label{ss:Syntacticcharacterisation}


In this part we offer two different syntactic characterisations of the family of POI revision operators. The first involves the following postulates:
\begin{tabbing}
\=BLAHI: \=\kill

\> \OmegaRevS{1} \> If $\neg A \not\in \bel{\Psi \ast A \vee B}$ and $A \not\in \bel{(\Psi \ast A) \ast B}$,  then $B \not\in \bel{(\Psi \ast B) \ast A}$ \\[0.1cm]

\> \OmegaRevS{2} \>  If $\neg A \not\in \bel{\Psi  \ast A \vee B}$ and $\neg A \in \bel{(\Psi \ast A) \ast B}$,  then $\neg B \in \bel{(\Psi \ast B) \ast A}$\\[0.1cm]

\> \OmegaRevS{3} \> If $\neg B \in \bel{\Psi \ast A \vee B}$ and $A \not\in \bel{(\Psi \ast A) \ast B}$,  then $\neg B \in \bel{(\Psi \ast B) \ast A}$ \\[-0.25em]	

\end{tabbing} 
\vspace{-1em}
These principles admit an interpretation in terms of the notions of  overruling and strictly overruling that we introduced in connection with \BetaRevPlusS{1} and \BetaRevPlusS{2}. Indeed, \OmegaRevS{1} and \OmegaRevS{2} stipulate conditions under which the obtaining of these relations entail that of their converses, while \OmegaRevS{3}~offers a condition that is sufficient for $B$'s overruling $A$ to entail $A$'s {\em strictly} overruling $B$. 

We remark that, for both lexicographic and restrained operators, it can be shown that the overrules and strictly overrules relations collapse into the same relation. Furthermore, for lexicographic revision, we have that $B$ overrules $A$ iff $A \wedge B$ is inconsistent (cf.\ the postulate \Rec~in Theorem \ref{beta1lex}), while, for restrained revision, $B$ overrules $A$ iff both $\neg A \in [\Psi * B]$ and $\neg B \in [\Psi *A]$ (i.e., iff $A$ and $B$ {\em counteract}, to use the terminology from \cite{booth2006admissible}). Clearly, in both cases, the overrules relation is symmetric, and so unrestricted versions of \OmegaRevS{1}-\OmegaRevS{3}~hold for these two sets of operators. 

With Proposition \ref{POIrepresentation} in mind, we now offer:

\begin{restatable}{prop}{SyntacticOmega}
\label{SyntacticOmega}
Given AGM, the following are equivalent: (a) \EquivalenceRevR, \CRevR{1}, \CRevR{2},  \IndRevR, \AlphaRevR{1}--\AlphaRevR{3}, and (b) \EquivalenceRevS, \CRevS{1}, \CRevS{2}, \BetaRevS{1}, \BetaRevS{2}, \OmegaRevS{1}--\OmegaRevS{3}.
\end{restatable}

\noindent While it employs some fairly accessible principles, this first characterisation  `bundles' the contribution of \IndRevR~into the principles  \OmegaRevS{1}--\OmegaRevS{3}. For this reason, we offer a second characterisation that separates out the contributions and maps each characteristic semantic principle onto a corresponding syntactic counterpart. Indeed, it turns out that the exact syntactic counterparts of \BetaRevR{3},  \BetaRevR{4}, and \AlphaRevR{3} are given as follows, where $\veebar$ denotes exclusive OR:

\begin{restatable}{prop}{AlphaSyntactic}
\label{AlphaSyntactic}
{\bf (a)} Given AGM, \BetaRevR{3}~is equivalent to 
\begin{tabbing}
\=BLAHI: \=\kill
\>\BetaRevS{3} \> If $B_2\notin \bel{\Psi \ast B_1}$, $B_1\rightarrow A\notin \bel{(\Psi \ast A) \ast B_2}$,  and $B_2\rightarrow \neg A\in \bel{\Psi \ast C}$, \\
\>\>then $B_2\wedge A\notin \bel{(\Psi \ast C) \ast B_1 \veebar B_2}$. \\[-0.25em]
\end{tabbing} 
\vspace{-1em}
{\bf (b)} Given AGM and \CRevR{4}, \BetaRevR{4} is equivalent to:
\begin{tabbing}
\=BLAHI: \=\kill
\> \BetaRevS{4}  \> If $B_2\notin \bel{\Psi \ast B_1}$, $B_1\wedge \neg A\in \bel{(\Psi \ast A) \ast B_2}$, and $B_2\rightarrow \neg A\in \bel{\Psi \ast C}$, \\
\>\>then $B_2\rightarrow \neg A\in \bel{(\Psi \ast C) \ast B_1 \veebar B_2}$. \\[-0.25em]
\end{tabbing} 
\vspace{-1em} 
{\bf (c)} Given AGM and \CRevR{3}, \AlphaRevR{3} is equivalent to:
\begin{tabbing}
\=BLAHI: \=\kill
\>\AlphaRevS{3} \> If $\neg B_2\in \bel{\Psi \ast B_1}$, $B_1\rightarrow A\notin \bel{(\Psi \ast A) \ast B_2}$,   and $B_2\rightarrow \neg A\in \bel{\Psi \ast C}$, \\
\>\>then $B_2\rightarrow \neg A\in \bel{(\Psi \ast C) \ast B_1 \veebar B_2}$. \\[-0.25em]
\end{tabbing} 
\vspace{-1em}
\end{restatable}

\noindent Since we already have syntactic counterparts for \BetaRevR{1} and \BetaRevR{2}, as well as \IndRevR, the above result completes a second syntactic characterisation of the POI family. This one-to-one correspondence between semantic and syntactic principles, however, comes at a cost, since we note that \AlphaRevS{1}--\AlphaRevS{3} are clearly much harder to interpret than \OmegaRevS{1}--\OmegaRevS{3}.

\section{Iterated versions of AGM era postulates}\label{s:Iteratedversions}


In this final section of the paper, we investigate various further properties of POI revision operators, discussing in the process an interesting issue that has  somewhat been neglected in the literature: the extension, to the iterated case, of the various AGM era postulates for revision. 

\subsection{Some postulates that are sound}

In \hyperref[s:Twoweakerprinciples]{`Two weaker principles'}, we briefly noted that \BetaRevR{1} and \BetaRevR{2} could each be reformulated as the conjunction of a pair of principles. We showed that these principles, which had not been discussed in the literature to date, are not  generally satisfied by admissible revision operators. It turns out, furthermore, that they are particularly noteworthy, since we can show that, in various combinations, they enable us to recover iterated generalisations of the following strong AGM postulates for revision and related well-known principles: 
\begin{tabbing}
\=BLAHI: \=\kill

\> \KRev{7} \> $\bel{\Psi\ast A\wedge C }\subseteq\textrm{Cn}(\bel{\Psi\ast A}\cup\{C\})$\\[0.1cm]

\> \DR \> $\bel{\Psi\ast A\vee C }\subseteq \bel{\Psi\ast A}\cup\bel{\Psi\ast C}$\\[0.1cm]

\> \DO \> $\bel{\Psi\ast A}\cap\bel{\Psi\ast C}\subseteq\bel{\Psi\ast A\vee C }$ \\[0.1cm]

\> \DI \> If $\neg A\notin \bel{\Psi\ast A\vee C }$, then $\bel{\Psi\ast A\vee C }\subseteq\bel{\Psi\ast A}$ \\[-0.25em]

\end{tabbing} 
\vspace{-1em}

\noindent \KRev{7}  is one of the two `supplementary' AGM postulates, and is also known as `Superexpansion'. `DR', `DO' and `DI' respectively abbreviate `Disjunctive Rationality', `Disjunctive Overlap' and `Disjunctive Inclusion'. As is well known in the literature, given  the other AGM postulates, \DR~is a consequence of the second supplementary postulate \KRev{8}, aka `Subexpansion', while \DO~is equivalent to \KRev{7} and \DI~to \KRev{8}.

The iterated generalisations that we recover are obtained by replacing all mentions of the belief states in the principles above by that of their corresponding revisions by a common sentence $B$ and making some minor adjustments. In each case, assuming Success and $\bel{\Psi \ast \top}=\bel{\Psi}$, setting $B= \top$ enables us to recover the non-iterated counterpart. We have:
\begin{tabbing}
\=BLAHI: \=\kill
\> \iKRev{7} \> $\bel{(\Psi\ast  A\wedge C)\ast B}\subseteq$ $\mathrm{Cn}(\bel{(\Psi \ast A)\ast B}\cup\{A\wedge C\})$  \\[0.1cm]

\> \iDR \> $\bel{(\Psi\ast A\vee C )\ast B}\subseteq$ $\bel{(\Psi\ast A)\ast B}\cup\bel{(\Psi\ast C)\ast B}$\\[0.1cm]

\> \iDO \> $\bel{(\Psi\ast A)\ast B}\cap\bel{(\Psi\ast C)\ast B}\subseteq$ $\bel{(\Psi\ast A\vee C )\ast B}$\\[0.1cm]

\> \iDI \> If $\neg A\notin \bel{(\Psi\ast A\vee C )\ast B}$,  then $\bel{(\Psi\ast A\vee C )\ast B}\subseteq\bel{(\Psi\ast A)\ast B}$  \\[-0.25em]

\end{tabbing} 
\vspace{-1em}
\noindent Although \iKRev{7} and \iDI~are, to the best of our knowledge, new to the literature, we note that \iDR~and \iDO~were already discussed and endorsed by Schlechta {\em et al} \shortcite{schlechta1996distance}. Our results are the following. 

\begin{restatable}{prop}{WPUSPUSyntactic}
\label{WPUSPUSyntactic}
In the presence of AGM,  \CRevR{1}  and \CRevR{2}, {\bf (a)}    \GammaRevR{1} and \GammaRevR{4} jointly entail \iDO~and {\bf (b)} \GammaRevR{2} and \GammaRevR{3} jointly entail \iDR.
\end{restatable}

\begin{restatable}{prop}{GammaOneTwoSyntactic}
\label{GammaOneTwoSyntactic}
Given AGM and \CRevR{1}, {\bf (a)} \GammaRevR{1} is equivalent to~\iDI~and {\bf (b)} \GammaRevR{2} is equivalent to \iKRev{7}.
\end{restatable}

\noindent Before moving on to the next subsection, a brief comment is in order regarding the well-known results, starting with the work of G\"ardenfors \shortcite{gard86},  which show that iterated versions of even the weak `basic' AGM postulates lead to triviality. How do these results square with our claims to recover iterated versions of various postulates that are substantially stronger? The answer lies in a difference in the approach to generalising these principles. In the literature surrounding the triviality results, the generalisations were obtained by replacing all references to belief sets by references to corresponding conditional belief sets. But this way of proceeding yields a significantly different set of generalisations to ours. To illustrate, consider the following `basic' AGM postulate, which is a weakening of \KRev{7} :

\begin{tabbing}
\=BLAHI: \=\kill
\> \KRev{3} \> $\bel{\Psi\ast A}\subseteq\textrm{Cn}(\bel{\Psi}\cup\{ A\})$\\[-0.25em]
\end{tabbing} 
\vspace{-1em}

\noindent In triviality result literature, however, its iterated counterpart is: $\cbel{\Psi\ast A}\subseteq\textrm{Cn}(\cbel{\Psi}\cup\{ A\})$. Etlin \shortcite{etlin2009} shows this principle to play badly with a pair of principles of conditional logic that he argues to be plausible. On our method for generating generalisations, however, we obtain the following weakening of our \iKRev{7}:

\begin{tabbing}
\=BLAHI: \=\kill
\> \iKRev{3} \> $\bel{(\Psi\ast A) \ast B}\subseteq\textrm{Cn}(\bel{\Psi \ast B}\cup\{ A\})$\\[-0.25em]
\end{tabbing} 
\vspace{-1em}

\noindent {\em This} principle is insufficiently strong to play the required role in the derivations of triviality and simply turns out to be a consequence of the DP postulates.\footnote{More specifically, it follows from \CRevR{1} and \CRevR{4}. Note that \CRevR{1} and \CRevR{3} enable us to also recover the following iterated version of `Preservation', which, given the other AGM postulates, is equivalent, in its non-iterated form, to \KRev{4}: \iPres~If $\neg A\notin \bel{\Psi \ast B}$, then $\bel{\Psi\ast B}\subseteq\bel{(\Psi\ast A)\ast B}$. If one adds \IndRevR~to these principles, one also recovers the corresponding iterated version of \KRev{4}: \iKRev{4}~If $\neg A\notin \bel{\Psi \ast B}$, then $\mathrm{Cn}(\bel{\Psi\ast B}\cup\{A\})\subseteq\bel{(\Psi\ast A)\ast B}$. We omit the proofs here, since these claims are not central to our discussion. \iKRev{3} and \iPres~are respective weakenings of \iKRev{7}~and \iDI. \iKRev{4}~is a weakening of the iterated version of Subexpansion, \iKRev{8}, discussed in the next section.}

\subsection{Some postulates that are {\em not} sound}

\noindent We have not recovered the iterated version of Subexpansion: 

\begin{tabbing}
\=BLAHI: \=\kill

\> \KRev{8} \> If $\neg C\notin \bel{\Psi\ast A}$, then $\textrm{Cn}(\bel{\Psi\ast A}\cup\{C\})\subseteq \bel{\Psi\ast A\wedge C}$\\[-0.25em]

\end{tabbing} 
\vspace{-1em}

\noindent which is given by:

\begin{tabbing}
\=BLAHI: \=\kill

\> \iKRev{8} \> If $\neg (A\wedge C)\notin \bel{(\Psi\ast A)\ast B}$, then $\textrm{Cn}(\bel{(\Psi\ast A)\ast B}\cup\{A\wedge C\})$\\ 
\>\>$\subseteq \bel{(\Psi\ast A\wedge C)\ast B}$  \\[-0.25em]

\end{tabbing} 
\vspace{-1em}

\noindent For this, we consider the following rather strong principle:
\begin{tabbing}
\=BLAHI: \=\kill
\>\IndRevPlusR \> If $x\in\mods{A}$, $y\in\mods{\neg A}$ and $x \preceq_{\Psi\ast A\vee  C} y$, then  $x \prec_{\Psi\ast A} y$ \\[-0.25em]
\end{tabbing} 
\vspace{-1em}

\noindent This principle strengthens the conjunction of \GammaRevR{1} and \GammaRevR{2} in much the same way that \IndRevR ~strengthens the conjunction of \CRevR{3} and \CRevR{4} (which, recall, are respective weakenings of \GammaRevR{1} and \GammaRevR{2}). Taken contrapositively, the principle inherits the weak antecedent of \GammaRevR{1} but the strong consequent of \GammaRevR{2}. Given $\preceq_{\Psi \ast \top} = \preceq_{\Psi}$, which follows from \CRevR{1}, \IndRevR~is recovered as the special case of \IndRevPlusR ~in which $C=\neg A$.
We now note:

\begin{restatable}{prop}{KrEightSem}
\label{prop:KrEightSem}
\iKRev{8}  is equivalent, given AGM and \CRevR{1}, to the conjunction of \GammaRevR{1} and \IndRevPlusR.
\end{restatable}

\noindent Where does our POI family stand with respect to this principle? Well we can establish the following:

\begin{restatable}{prop}{IndPlusSoundness}
\label{IndPlusSoundness}
\IndRevPlusR ~is satisfied by both lexicographic and restrained revision operators.
\end{restatable}

\noindent  Since lexicographic and restrained revision operators satisfy \GammaRevPlusR{1}, this establishes, that they satisfy \iKRev{8}. This is interesting, since it shows, not only that the principle is consistent with our previous constraints, but that adding it to these does not yield the kind of `reductionism' that has been argued to be objectionable.  However, it remains the case that 

\begin{restatable}{prop}{IndPlusNotSoundForPOIs}
\label{IndPlusNotSoundForPOIs}
\IndRevPlusR ~is not generally satisfied by POI revision operators.
\end{restatable}

\noindent In fact, a weaker property than this one fails to hold across the family. Indeed, \IndRevPlusR~generalises the following {\em Separation} property, discussed by Booth \& Meyer \shortcite{booth2006admissible} under the name of `UR', which is the special case of \IndRevPlusR~in which $C = A$:

\begin{tabbing}
\=BLAHI: \=\kill
\> \SepRevR \> If $x\in\mods{A}$ and $y\in\mods{\neg A}$, then $x \prec_{\Psi\ast A} y$ or  $y \prec_{\Psi\ast A} x$ \\[-0.25em]
\end{tabbing} 
\vspace{-1em}

\noindent This condition can be captured by a `{\em Non-Flush}' constraint on the POI assignment, which states that it is never the case that two intervals line up flush, in the sense that $x^{+}\sim y^{-}$. This condition is not satisfied in general by POI assignments. Indeed, consider the following POI assignment to $W=\{x,y,z\}$: $x^{+} <  y^{+} < x^{-} < y^{-} \sim z^{+} < z^{-}$. Non-Flush fails, with the result that so too does \SepRevR~and hence \IndRevPlusR, since $y\sim_{\Psi \ast x \vee z} z$. This establishes Proposition \ref{IndPlusNotSoundForPOIs}.

At this point, a natural question arises: Why has the narrower family of POI revision operators satisfying \SepRevR, or indeed, \IndRevPlusR, not made a more central appearance in the present paper? The answer to this is that \SepRevR~remains in our view an extremely strong property. This becomes most apparent when one considers its syntactic counterpart:

\begin{tabbing}
\=BLAHI: \=\kill
\> \SepRevS \> Either $\neg A\in\bel{(\Psi \ast A) \ast B}$ or $A\in\bel{(\Psi \ast A) \ast B}$ \\[-0.25em]
\end{tabbing} 
\vspace{-1em}

\noindent This principle states that, once one has revised one's beliefs by a certain sentence, one will remain opinionated as to whether or not that sentence is true upon any further single revision. But this seems too  strong: let $A$ be any sentence and $B$ be the sentence `The Oracle says that it might not be the case that $A$'. Plausibly $A, \neg A\notin\bel{(\Psi \ast A) \ast B}$.

Also of interest is the iterated version of `Disjunctive Factoring',  which is equivalent to the conjunction of \KRev{7} and \KRev{8}, in the presence of the other AGM postulates:\footnote{This condition is typically stated in weaker terms, as: $\bel{\Psi\ast A\vee C }$ is equal to either $\bel{\Psi\ast A}$, $\bel{\Psi\ast C}$, or $\bel{\Psi\ast A}\cap \bel{\Psi\ast C}$. However, the equivalence that is proven is in fact with the stronger principle. See \cite[Proposition 3.16]{GardFlux}, where the proof is credited to Hans Rott.} 
\begin{tabbing}
\=BLAHBI: \= BLA \=\kill
\>\DF \> {(i)} \> If $\neg C\in  \bel{\Psi\ast A\vee C}$, then  $\bel{\Psi\ast A\vee C}=\bel{\Psi\ast A}$ \\[0.1cm]
\> \> {(ii)} \> If $\neg A,\neg C\notin\bel{\Psi\ast A\vee C}$, then $\bel{\Psi\ast A\vee C}=\bel{\Psi\ast A}\cap \bel{\Psi\ast C}$ \\[0.1cm]
\> \> {(iii)} \> If $\neg A\in\bel{\Psi\ast A\vee C}$, then $\bel{\Psi\ast A\vee C}=\bel{\Psi\ast C}$\\[-0.25em]
\end{tabbing} 
\vspace{-1em}

%
%

\noindent The iterated version is given by:
\begin{tabbing}
\=BLAHBI: \= BLA \=\kill
\>\iDF \> {(i)} \> If $\neg C\in  \bel{(\Psi\ast A\vee C)\ast B}$, then $\bel{(\Psi\ast A\vee C)\ast B}$\\
\>\>\>$=\bel{(\Psi\ast A)\ast B}$ \\[0.1cm]
\> \> {(ii)} \> If $\neg A,\neg C\notin\bel{(\Psi\ast A\vee C)\ast B}$, then $\bel{(\Psi\ast A\vee C)\ast B}$\\
\>\>\>$=\bel{(\Psi\ast A)\ast B}$ $\cap \bel{(\Psi\ast C)\ast B}$ \\[0.1cm]
\> \> {(iii)} \> If $\neg A\in\bel{(\Psi\ast A\vee C)\ast B}$, then $\bel{(\Psi\ast A\vee C)\ast B}$\\
\>\>\>$=\bel{(\Psi\ast C)\ast B}$\\[-0.25em]
\end{tabbing} 
\vspace{-1em}

\noindent \iDF (ii) is entailed by the combination of \iDO, for the right-to-left direction, and \iDI, for the left-to-right direction, both of which we have established to be sound for POI revision operators. Regarding \iDF (i):

 \begin{restatable}{prop}{DFTwoLtoRSemantic}
\label{DFTwoLtoRSemantic}
The semantic counterparts of the right-to-left and left-to-right directions of \iDF {\em (i)} are respectively: 
\begin{tabbing}
\=BLAHI: \=\kill
\>\GammaRevR{5} \> If $x, y \in \mods{\neg A}$ and $y \preceq_{\Psi \ast A \vee C} x$, then $y \preceq_{\Psi \ast C} x$  \\[0.1em] 
\>\GammaRevR{6} \> If $y \in \mods{\neg A}$, and $y \prec_{\Psi \ast A \vee C} x$, then $y \prec_{\Psi \ast C} x$. \\[-0.25em]
\end{tabbing} 
\vspace{-1em}
\end{restatable}

\noindent We note that the second of these two principles, in conjunction with \GammaRevR{2}, obviously gives us \BetaRevPlusR{2}. However, due to the requirement that $x \in \mods{\neg A}$ in the antecedent of the second principle, the latter does not give us \BetaRevPlusR{1}, in conjunction with \GammaRevR{1}. Where do they stand in relation to our family of operators? The answer is the following:

 \begin{restatable}{prop}{DFTwoLtoRNotSoundPOI}
\label{DFTwoLtoRNotSoundPOI}
Neither \GammaRevR{5} nor \GammaRevR{6} are generally satisfied by POI revision operators.
\end{restatable}

\noindent However:

 \begin{restatable}{prop}{DFTwoLtoRSoundLex}
\label{DFTwoLtoRSoundLex}
Both \GammaRevR{5} and \GammaRevR{6} are satisfied by lexicographic revision operators.
\end{restatable}

\noindent This establishes that \iDF~is satisfied by lexicographic revision operators, since we have individually shown that they satisfy all the component principles. 


\section{Conclusions and further work}\label{s:Conclusions}


This paper has investigated a significant, yet comparatively  restrained, strengthening of the seminal framework introduced two decades ago by Darwiche and Pearl. Unlike the majority of existing models of iterated revision, the proposal falls short of identifying belief states with simple total preorders over worlds. Indeed, it incorporates further structure into these, in the form of proper ordinal intervals.\footnote{We are not the only ones to have proposed an enrichment of belief states beyond mere TPOs or equivalent structures. One well-known case in point is Spohn's identification of states with `{\em ranking functions}', aka `OCFs' \cite{Spohn1988-SPOOCF,Spohn2012-SPOTLO}. However, Spohn does not acknowledge the concept of revision simpliciter that we are studying. Rather, he considers a parameterised {\em family} of revision-like  functions. This additional parameter very much complicates the translation of our principles into his framework. In a rather different vein, Konieczny \& P\'erez \shortcite{doi:10.1080/11663081.2000.10511003} identify states with {\em histories of input sentences} (see also \cite{Lehmann:1995:BRR:1643031.1643098} and \cite{DBLP:conf/kr/DelgrandeDL06} for related work). However, it turns out that if \CRevS{2}--a feature of POI revision--is imposed, then the class of operators that they study narrows down to lexicographic revision alone.
}
This is achieved by combining Booth \& Meyer's framework for non-prioritised revision with a `naturalisation' step, in a move that bears some similarities to the definition, via the Levi Identity, of single-step revision as a contraction followed by an expansion. The resulting family of POI revision operators, which is a sub-family of the so-called `admissible' family, has been characterised both semantically and syntactically. It has also  been shown that POI revision operators are distinctive, within the class of admissible ones, in satisfying iterated counterparts of many (albeit not all) classic AGM era postulates. 

In future work, we first plan to consider the consequences of relaxing  \IndRevS. This condition fails for a more general family of what one could call `{\em basic ordinal interval (BOI)}' revision operators. These operators, which include natural revision operators, are naturalisations of non-prioritised operators based on ordinal interval assignments that satisfy, not \Leq{2}, but the weaker requirement that $x^{+} \leq x^{-}$. As it turns out, our proof of the soundness of \AlphaRevR{1}--\AlphaRevR{3} with respect to POI operators carries over here, leaving us in a strong position to provide a characterisation for this more general family. 
Secondly, as Figure \ref{fig:setup} reminds us, the constraints that we have discussed impose few constraints on the result of more than two  iterations of the revision operation. While the structure associated with belief states currently determines the posterior {\em TPO}, nothing has been said regarding the nature of the posterior {\em POI}.

\bibliographystyle{aaai}
\bibliography{ELIR-Long}

%



\section*{Appendix}\label{s:Appendix}


\CircPropertyPreservation*

\begin{newproof}
Recall that $\mathbb{N}(\ast, \circ)$ iff: $x \preceq_{\Psi \ast A} y$ iff either (i) $x \in \min(\preceq_{\Psi\circ A}, \mods{A})$, or (ii) $x, y \notin \min(\preceq_{\Psi\circ A}, \mods{A})$ and $x \preceq_{\Psi\circ A} y$.
\begin{itemize}

\item[{\bf (i)}] {\bf Preservation of \CRevR{1}:} Assume that $x,y \in \mods{A}$ and that $x \preceq_{\Psi \circ A} y$ iff  $x \preceq_\Psi y$. We show that $x \preceq_{\Psi \circ A} y$ iff  $x \preceq_{\Psi \ast A} y$. 

From $x \preceq_{\Psi \ast A} y$ to $x \preceq_{\Psi \circ A} y$: Trivial.
 
From $x \preceq_{\Psi \circ A} y$ to $x \preceq_{\Psi \ast A} y$: Assume $x \preceq_{\Psi \circ A} y$. We just need to show that, if  $x \notin \min(\preceq_{\Psi\circ A}, \mods{A})$, then $y \notin \min(\preceq_{\Psi\circ A}, \mods{A})$. This follows immediately from $x \preceq_{\Psi \circ A} y$ and  $x \in \mods{A}$.

\item[{\bf (ii)}] {\bf Preservation of \CRevR{2}:} Assume that $x,y \in \mods{A}$ and that $x \preceq_{\Psi \circ A} y$ iff  $x \preceq_\Psi y$. We show that $x \preceq_{\Psi \circ A} y$ iff  $x \preceq_{\Psi \ast A} y$.  

From $x \preceq_{\Psi \ast A} y$ to $x \preceq_{\Psi \circ A} y$: Trivial.
 
From $x \preceq_{\Psi \circ A} y$ to $x \preceq_{\Psi \ast A} y$: Assume $x \preceq_{\Psi \circ A} y$. We just need to show that, if  $x \notin \min(\preceq_{\Psi\circ A}, \mods{A})$, then $y \notin \min(\preceq_{\Psi\circ A}, \mods{A})$. This follows immediately from $y \in \mods{\neg A}$.

\item[{\bf (iii)}] {\bf Preservation of \IndRevR:} Assume that $x \in \mods{A}$, $y \in \mods{\neg A}$ and that, if $x \preceq_{\Psi} y$ then $x \prec_{\Psi \circ A} y$. We show that, if $x \prec_{\Psi \circ A} y$, then $x \prec_{\Psi \ast A} y$. From the definition of naturalisation, we have: if $\mathbb{N}(\ast, \circ)$, then $x \prec_{\Psi \ast A} y$ iff either (i) $x \in \min(\preceq_{\Psi\circ A}, \mods{A})$ and $y \notin \min(\preceq_{\Psi\circ A}, \mods{A})$, or (ii) $x, y \notin \min(\preceq_{\Psi\circ A}, \mods{A})$ and $x \prec_{\Psi\circ A} y$. So, given $x \prec_{\Psi \circ A} y$, we just need to show that, if $x \in \min(\preceq_{\Psi\circ A}, \mods{A})$, then $y \notin \min(\preceq_{\Psi\circ A}, \mods{A})$. This follows immediately from $y \in \mods{\neg A}$. 

\item[{\bf (iv)}] {\bf Preservation of \EquivalenceRevR:} Assume $A\equiv B$. We want to show $\preceq_{\Psi\ast A}=\preceq_{\Psi\ast B}$. So assume for contradiction that there exist $x$ and $y$ such that $x \preceq_{\Psi\ast A} y$ but $y \prec_{\Psi\ast B} x$ (the other case is analogous). 

From $x \preceq_{\Psi\ast A} y$ we have: (1) $x \in \min(\preceq_{\Psi\circ A}, \mods{A})$, or (2) $x, y \notin \min(\preceq_{\Psi\circ A}, \mods{A})$ and $x \preceq_{\Psi\circ A} y$. 

From $y \prec_{\Psi\ast A} x$, we obtain:  (3) $y \in \min(\preceq_{\Psi\circ B}, \mods{B})$ and $x \notin \min(\preceq_{\Psi\circ B}, \mods{B})$, or (4) $x, y \notin \min(\preceq_{\Psi\circ B}, \mods{B})$ and $y \prec_{\Psi\circ B} y$.

From $A\equiv B$  and the fact that $\circ$ satisfies \EquivalenceRevR, we have $\preceq_{\Psi\circ A}=\preceq_{\Psi\circ B}$ and $\mods{A}=\mods{B}$. Given this, it is easy to see that neither (1) nor (2) is consistent with either (3) or (4).   \qed

\end{itemize}

\end{newproof}

\vspace{1em}


\BetaOneTwoSyntactic*

\begin{newproof} 

\begin{itemize}

\item[{\bf (a)}]
\begin{itemize}

\item[{\bf (i)}] {\bf From \BetaRevR{1} to \BetaRevS{1}:} Assume $A \notin \bel{(\Psi \ast A) \ast B}$, $B \rightarrow \neg A \in \bel{\Psi \ast C}$ and, for contradiction, $A \in \bel{(\Psi \ast C) \ast B}$. From $A \in \bel{(\Psi \ast C) \ast B}$, it follows that $\min(\preceq_{\Psi \ast C}, \mods{B})\subseteq \mods{A}$. Now consider an arbitrary $x \in \min(\preceq_{\Psi \ast C}, \mods{B})$. Since $x \in \mods{A \wedge B}$, it follows from $\neg(A \wedge B) \in \bel{\Psi \ast C}$ and Success that $x \notin \min(\preceq, \mods{C})$. From $A \notin \bel{(\Psi \ast A) \ast B}$, there exists $y$ such that  $y \in \mods{\neg A} \cap \min(\preceq_{\Psi \ast A}, \mods{B})$. Given $x \in \mods{B}$, we furthermore have $y \preceq_{\Psi \ast A} x$. By \BetaRevR{1}, we then recover $y \preceq_{\Psi \ast C} x$. Since $y \in \mods{B}$, $x \in \min(\preceq_{\Psi \ast C}, \mods{B})$ and $y \in \mods{\neg A}$,
 this contradicts $\min(\preceq_{\Psi \ast C}, \mods{B})\subseteq \mods{A}$. Hence $A \notin \bel{(\Psi \ast C) \ast B}$, as required.

\item[{\bf (ii)}] {\bf From \BetaRevS{1} to \BetaRevR{1}:} Assume $x \notin \min(\preceq, \mods{C})$, $x \in \mods{A}$, $y \in \mods{\neg A}$ and $y \preceq_{\Psi \ast A} x$. Assume for contradiction that $\neg(A \wedge (x \vee y)) \notin \bel{\Psi \ast C}$. Then there exists $w$ in $\min(\preceq, \mods{C}) \cap \mods{A \wedge  (x \vee y)}$. But $\mods{A \wedge (x \vee y)} = \{x\}$, since $x \in \mods{A}$ and $y \in \mods{\neg A}$, and we have assumed $x \notin  \min(\preceq, \mods{C})$. Contradiction. So $\neg(A \wedge (x \vee y)) \in \bel{\Psi \ast C}$. From $y \in \mods{\neg A}$ and $y \preceq_{\Psi \ast A} x$, it follows that $A \notin \bel{(\Psi \ast A) \ast x \vee y}$. By \BetaRevS{1}, we then recover $A \notin \bel{(\Psi \ast C) \ast x \vee y}$ and hence, since $x \in \mods{A}$ and $y \in \mods{\neg A}$,  $y \preceq_{\Psi\ast C} x$, as required.
\end{itemize}

\item[{\bf (b)}]
\begin{itemize}

\item[{\bf (i)}] {\bf From \BetaRevR{2} to \BetaRevS{2}:} Assume $\neg A \in \bel{(\Psi \ast A) \ast B}$, $B \rightarrow \neg A \in \bel{\Psi \ast C}$ and, for contradiction, $\neg A \notin \bel{(\Psi \ast C) \ast B}$. From $\neg A \notin \bel{(\Psi \ast C) \ast B}$, there exists $x$ such that  $x \in \mods{A} \cap \min(\preceq_{\Psi \ast C}, \mods{B})$. Since $x \in \mods{A \wedge B}$, it follows from $\neg(A \wedge B) \in \bel{\Psi \ast C}$ and Success that $x \notin \min(\preceq, \mods{C})$. From $\neg A \in \bel{(\Psi \ast A) \ast B}$, there exists $y$ such that  $y \in \mods{\neg A} \cap \min(\preceq_{\Psi \ast A}, \mods{B})$. Given $x \in \mods{B}$, we furthermore have $y \prec_{\Psi \ast A} x$. By \BetaRevR{2}, we then recover $y \prec_{\Psi \ast C} x$. Since $y \in \mods{B}$, this contradicts $x \in \min(\preceq_{\Psi \ast C}, \mods{B})$. Hence $\neg A \in \bel{(\Psi \ast C) \ast B}$, as required. 

\item[{\bf (ii)}] {\bf From \BetaRevS{2} to \BetaRevR{2}:} Assume $x \notin \min(\preceq, \mods{C})$, $x \in \mods{A}$, $y \in \mods{\neg A}$ and $y \prec_{\Psi \ast A} x$. Assume for contradiction that $\neg(A \wedge (x \vee y)) \notin \bel{\Psi \ast C}$. Then there exists $w$ in $\min(\preceq, \mods{C}) \cap \mods{A \wedge  (x \vee y)}$. But $\mods{A \wedge (x \vee y)} = \{x\}$, since $x \in \mods{A}$ and $y \in \mods{\neg A}$, and we have assumed $x \notin  \min(\preceq, \mods{C})$. Contradiction. So $\neg(A \wedge (x \vee y)) \in \bel{\Psi \ast C}$. From $y \in \mods{\neg A}$ and $y \prec_{\Psi \ast A} x$, it follows that $\neg A \in \bel{(\Psi \ast A) \ast x \vee y}$. By \BetaRevS{2}, we then recover $\neg A \in \bel{(\Psi \ast C) \ast x \vee y}$ and hence, since $x \in \mods{A}$ and $y \in \mods{\neg A}$,  $y \prec_{\Psi\ast C} x$, as required. \qed
\end{itemize}

\end{itemize}
\end{newproof}

\vspace{1em}


\BetaGammaEquivalence*

\begin{newproof} 

\begin{itemize} 

\item[{\bf (a)}] 
\begin{itemize} 

\item[{\bf (i)}] {\bf From \GammaRevR{1} and \GammaRevR{3}  to \BetaRevR{1}:} From \CRevR{2}  and \CRevR{4}, we obtain:
\begin{tabbing}
\=BLAHII: \=\kill
\> (1) \> If $x \notin \min(\preceq, \mods{C})$, $x \in \mods{A \vee C}$, $y \in \mods{C}$  and $y \preceq_{\Psi \ast A \vee C} x$, \\
\> \> then $y \preceq_{\Psi \ast C} x$ \\[-0.25em]
\end{tabbing} 
\vspace{-1em}
Indeed, from $x \in \mods{A \vee C}$, $y \in \mods{C} \subseteq \mods{A \vee C}$  and $y \preceq_{\Psi \ast A \vee C} x$, \CRevR{2}  gives us $y \preceq_{\Psi} x$. Given $y \in \mods{C}$, if we assume  $x \in \mods{C}$, then $y \preceq_{\Psi \ast C} x$ follows from \CRevR{2}. If we assume instead that $x \in \mods{\neg C}$, the same conclusion follows from \CRevR{4}. The conjunction of \GammaRevR{3} and (1) then gives us:
\begin{tabbing}
\=BLAHII: \=\kill
\> (2) \> If $x \notin \min(\preceq, \mods{C})$, $x \in \mods{A \vee C}$, $y \in \mods{\neg A \vee C}$ and \\
\> \> $y \preceq_{\Psi \ast A \vee C} x$, then $y \preceq_{\Psi \ast C} x$ \\[-0.25em]
\end{tabbing} 
\vspace{-1em}
We now show that (2) and \GammaRevR{1} entail \BetaRevR{1}: Assume $x \in \mods{A}$, $y \in \mods{\neg A}$, $x \notin \min(\preceq, \mods{C})$ and $y \preceq_{\Psi \ast A} x$. From $x \in \mods{A}$, $y \in \mods{\neg A}$, and $y \preceq_{\Psi \ast A} x$, \GammaRevR{1} gives us $y \preceq_{\Psi \ast A \vee C}x$. From $x \in \mods{A}$ and $y \in \mods{\neg A}$, we recover $x \in \mods{A \vee C}$ and $y \in \mods{\neg A \vee C}$. From $x \notin \min(\preceq, \mods{C})$, $x \in \mods{A \vee C}$, $y \in \mods{\neg A \vee C}$ and $y \preceq_{\Psi \ast A \vee C} x$, (2) gives us $y \preceq_{\Psi \ast C} x$, as required.

\item[{\bf (ii)}] {\bf From \BetaRevR{1} to \GammaRevR{1} and \GammaRevR{3}:} To get from \BetaRevR{1} to \GammaRevR{3}, simply substitute $A \vee C$ for $A$ in \BetaRevR{1}. To get from \BetaRevR{1} to \GammaRevR{1},  assume $x \in \mods{A}$, $y \in  \mods{\neg A}$ and $y \preceq_{\Psi \ast A} x$. From $y \preceq_{\Psi \ast A} x$ and $y \in  \mods{\neg A}$, it follows by Success that $x \notin \min(\preceq, \mods{A})$ and hence by $x \in \mods{A}$ that $x \notin \min(\preceq, \mods{A \vee C})$. By \BetaRevR{1}, we then recover $y \preceq_{\Psi \ast A \vee C} x$, as required.

\end{itemize}
\item[{\bf (b)}]
\begin{itemize} 

\item[{\bf (i)}] {\bf From \GammaRevR{2} and \GammaRevR{4}  to \BetaRevR{2}:} From \CRevR{1}  and \CRevR{3}, we obtain:
\begin{tabbing}
\=BLAHII: \=\kill
\> (1) \> If $x \notin \min(\preceq, \mods{C})$, $x \in \mods{A \vee C}$,  $y \in \mods{C}$  and $y \prec_{\Psi \ast A \vee C} x$, \\
\> \> then $y \prec_{\Psi \ast C} x$ \\[-0.25em]
\end{tabbing} 
\vspace{-1em}
Indeed, from $x \in \mods{A \vee C}$, $y \in \mods{C} \subseteq \mods{A \vee C}$  and $y \prec_{\Psi \ast A \vee C} x$, \CRevR{1}  gives us $y \prec_{\Psi} x$. Given $y \in \mods{C}$, if we assume  $x \in \mods{C}$, then $y \prec_{\Psi \ast C} x$ follows from \CRevR{1}. If we assume instead that $x \in \mods{\neg C}$, the same conclusion follows from \CRevR{3}. The conjunction of \GammaRevR{4}~and (1) then gives us:
\begin{tabbing}
\=BLAHII: \=\kill
\> (2) \> If $x \notin \min(\preceq, \mods{C})$, $x \in \mods{A \vee C}$,  $y \in \mods{\neg A \vee C}$ and  \\
\> \> $y \prec_{\Psi \ast A \vee C} x$, then $y \prec_{\Psi \ast C} x$ \\[-0.25em]
\end{tabbing} 
\vspace{-1em}
We now show that (2) and \GammaRevR{2}  entail \BetaRevR{2}: Assume $x \in \mods{A}$, $y \in \mods{\neg A}$, $x \notin \min(\preceq, \mods{C})$ and $y \prec_{\Psi \ast A} x$. From $x \in \mods{A}$, $y \in \mods{\neg A}$, and $y \prec_{\Psi \ast A} x$, \GammaRevR{2}  gives us $y \prec_{\Psi \ast A \vee C}x$. From $x \in \mods{A}$ and $y \in \mods{\neg A}$, we recover $x \in \mods{A \vee C}$ and $y \in \mods{\neg A \vee C}$. From $x \notin \min(\preceq, \mods{C})$, $x \in \mods{A \vee C}$, $y \in \mods{\neg A \vee C}$ and $y \prec_{\Psi \ast A \vee C} x$, (2) gives us $y \prec_{\Psi \ast C} x$, as required.

\item[{\bf (ii)}] {\bf From \BetaRevR{2} to the conjunction of \GammaRevR{2} and \GammaRevR{4}:} To get from \BetaRevR{2} to \GammaRevR{4}, simply substitute $A \vee C$ for $A$ in \BetaRevR{2}. To get from \BetaRevR{2} to \GammaRevR{2},   assume $x \in \mods{A}$, $y \in  \mods{\neg A}$ and $y \prec_{\Psi \ast A} x$. From $y \prec_{\Psi \ast A} x$, it follows by Success that $x \notin \min(\preceq, \mods{A})$ and hence by $x \in \mods{A}$ that $x \notin \min(\preceq, \mods{A \vee C})$. By \BetaRevR{2}, we then recover $y \prec_{\Psi \ast A \vee C} x$, as required. \qed

\end{itemize}
\end{itemize}

\end{newproof}

\vspace{1em}


\wSriTwoSemanticComment*

\begin{newproof}

\begin{itemize}

\item[{\bf (a)}]  {\bf Regarding \GammaRevR{1} and \GammaRevR{2}:}   Consider the countermodel below. The boxes represent states and associated TPOs. The pairs of characters represent worlds, with $AC\in\mods{A\wedge C}$, $\overline{A}\overline{C}\in\mods{\neg A\wedge \neg C}$, $A\overline{C}\in\mods{A\wedge \neg C}$ and $\overline{A}C\in\mods{\neg A\wedge C}$. The worlds are ordered from top to bottom by decreasing order of `plausibility'. It is easily verified that \CRevR{1},  \CRevR{2}~and \IndRevR~are all satisfied. However, both \GammaRevR{1}~and \GammaRevR{2}~are violated, since $\overline{A}\overline{C}\prec_{\Psi\ast A} A\overline{C}$ but $A\overline{C}\prec_{\Psi\ast A\vee C} \overline{A}\overline{C}$.

\end{itemize}

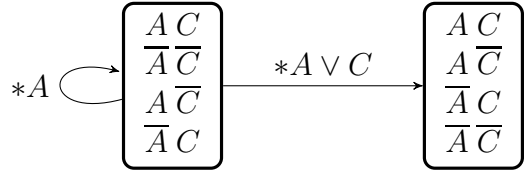
\bigskip

\begin{centering}
\begin{tikzpicture}[->,>=stealth']
\scalebox{1}{%
 \node[state] (K) 
 {
\begin{tabular}{c}
$A~C$\\
$\overline{A}~\overline{C}$\\
$A~\overline{C}$\\
$\overline{A}~C$\\
\end{tabular}
};

 \node[state,       
 node distance=4cm,     
 right of=K,        
 yshift=+0cm] (K*AVC)    
 {
\begin{tabular}{c}
$A~C$\\
$A~\overline{C}$\\
$\overline{A}~C$\\
$\overline{A}~\overline{C}$\\
\end{tabular}
 };

 \path (K) edge  node[anchor=north,above]
                   {
                   $\ast A\vee C$
                   } (K*AVC)
;

 \path (K)  	edge[loop left]    node[anchor=north,left]{$\ast A$} (K)

;
}
\end{tikzpicture}

\end{centering}

\begin{itemize}

\item[{\bf (b)}]  {\bf Regarding \GammaRevR{3}~and \GammaRevR{4}:}   Consider the countermodel below. It is easily verified that \CRevR{1},  \CRevR{2}~and \IndRevR~are all satisfied. However, both \GammaRevR{3}~and \GammaRevR{4}~are violated, since, although $\overline{A}C\notin\min{\preceq_{\Psi}, \mods{C}}$, we have $\overline{A}\overline{C}\prec_{\Psi\ast A\vee C} \overline{A}C$ but $\overline{A}C\prec_{\Psi\ast C} \overline{A}\overline{C}$.

\end{itemize}

\bigskip

\begin{centering}
\begin{tikzpicture}[->,>=stealth']
\scalebox{1}{%
 \node[state] (K) 
 {
\begin{tabular}{c}
$A~C$\\
$\overline{A}~\overline{C}$\\
$\overline{A}~C$\\
$A~\overline{C}$\\
\end{tabular}
};

 \node[state,       
 node distance=4cm,     
 right of=K,        
 yshift=+0cm] (K*AVC)    
 {
\begin{tabular}{c}
$A~C$\\
$\overline{A}~C$\\
$\overline{A}~\overline{C}$\\
$A~\overline{C}$\\
\end{tabular}
 };

 \path (K) edge  node[anchor=north,above]
                   {
                   $\ast C$
                   } (K*AVC)
;

 \path (K)  	edge[loop left]    node[anchor=north,left]{$\ast A\vee C$} (K)

;
}
\end{tikzpicture}

\end{centering}

\qed

\end{newproof}

\vspace{1em}


\SoundnessAlphas*

\begin{newproof} We first note that, for POI operators, $y \prec_{\Psi \ast A} x$ iff 
\begin{itemize}

\item[(1)] $y \in \min(\preceq, \mods{A})$ and $x \notin \min(\preceq, \mods{A})$, or

\item[(2)] $x, y \notin \min(\preceq, \mods{A})$ and $r_{A}(y) < r_{A}(x)$

\end{itemize}
We also have $z \prec_{\Psi \ast C} x$ iff either
\begin{itemize}

\item[(3)] $z \in \min(\preceq, \mods{C})$ and $x \notin \min(\preceq, \mods{C})$, or

\item[(4)] $z, x \notin \min(\preceq, \mods{C})$ and $r_{C}(z) < r_{C}(x)$

\end{itemize}
as well as $y \preceq_{\Psi \ast A} x$ iff 
\begin{itemize}

\item[(5)] $y \in \min(\preceq, \mods{A})$, or

\item[(6)] $x, y \notin \min(\preceq, \mods{A})$ and $r_{A}(y) \leq r_{A}(x)$

\end{itemize}
and finally $z \preceq_{\Psi \ast C} x$ iff either
\begin{itemize}

\item[(7)] $z \in \min(\preceq, \mods{C})$, or

\item[(8)] $z, x \notin \min(\preceq, \mods{C})$ and $r_{C}(z) \leq r_{C}(x)$

\end{itemize}
With this in hand, we prove the soundness of each principle in turn:
\begin{itemize}

\item[{\bf (a)}] {\bf Regarding \AlphaRevR{1}:} Assume $x \notin \min(\preceq, \mods{C})$, $x \in \mods{A}$, $y \in \mods{\neg A}$, $z \preceq y$ and $y \preceq_{\Psi\ast A}x$. Since $y \in \mods{\neg A}$, we have $y \notin \min(\preceq, \mods{A})$, placing us in case (6). So $r_{A}(y) \leq r_{A}(x)$. Since $x \in \mods{A}$, we have $r_{A}(x) = x^{+}$ and, since $y \in \mods{\neg A}$, it follows that $r_{A}(y) = y^{-}$. So $y^{-} \leq x^{+}$. Furthermore, since $z \preceq y$, we have $z^{-} \leq y^{-}$. Hence $z^{-} \leq x^{+}$. Since $z^{+} \leq z^{-}$ and $x^{+} \leq x^{-}$, it then follows that $r_{C}(z) \leq r_{C}(x)$. Since we have also assumed $x \notin \min(\preceq, \mods{C})$, if $z \in \min(\preceq, \mods{C})$, we are in case (7) and if  $z \notin \min(\preceq, \mods{C})$, we are in case (8). Either way, we have $z \preceq_{\Psi \ast C} x$, as required. 

\item[{\bf (b)}] {\bf Regarding \AlphaRevR{2}:}  Assume $x \notin \min(\preceq, \mods{C})$, $x \in \mods{A}$, $y \in \mods{\neg A}$, $z \preceq y$ and $y \prec_{\Psi\ast A}x$. Since $y \in \mods{\neg A}$, we have $y \notin \min(\preceq, \mods{A})$, placing us in case (2). So $r_{A}(y) < r_{A}(x)$. Since $x \in \mods{A}$, we have $r_{A}(x) = x^{+}$ and, since $y \in \mods{\neg A}$, it follows that $r_{A}(y) = y^{-}$. So $y^{-} < x^{+}$. Furthermore, since $z \preceq y$, we have $z^{-} \leq y^{-}$. Hence $z^{-} < x^{+}$. Since $z^{+} \leq z^{-}$ and $x^{+} \leq x^{-}$, it then follows that $r_{C}(z) < r_{C}(x)$. Since we have also assumed $x \notin \min(\preceq, \mods{C})$, if $z \in \min(\preceq, \mods{C})$, we are in case (3) and if  $z \notin \min(\preceq, \mods{C})$, we are in case (4). Either way, we have $z \prec_{\Psi \ast C} x$, as required. 

\item[{\bf (c)}] {\bf Regarding \AlphaRevR{3}:}  Assume $x \notin \min(\preceq, \mods{C})$, $x \in \mods{A}$, $y \in \mods{\neg A}$, $z \prec y$ and $y \preceq_{\Psi\ast A}x$. Since $y \in \mods{\neg A}$, we have $y \notin \min(\preceq, \mods{A})$, placing us in case (6). So $r_{A}(y) \leq r_{A}(x)$. Since $x \in \mods{A}$, we have $r_{A}(x) = x^{+}$ and, since $y \in \mods{\neg A}$, it follows that $r_{A}(y) = y^{-}$. So $y^{-} \leq x^{+}$. Furthermore, since $z \prec y$, we have $z^{-} <  y^{-}$. Hence $z^{-} < x^{+}$. Since $z^{+} \leq z^{-}$ and $x^{+} \leq x^{-}$, it then follows that $r_{C}(z) < r_{C}(x)$. Since we have also assumed $x \notin \min(\preceq, \mods{C})$, if $z \in \min(\preceq, \mods{C})$, we are in case (3) and if  $z \notin \min(\preceq, \mods{C})$, we are in case (4). Either way, we have $z \prec_{\Psi \ast C} x$, as required. \qed
\end{itemize}

\end{newproof}

\vspace{1em}


\POIrepresentation*
\begin{newproof} In conjunction with the results of Booth \& Meyer regarding non-prioritised POI revision operators, Propositions \ref{CircPropertyPreservation} and  \ref{SoundnessAlphas} establish the left-to-right direction of the claim. We simply need to establish the other direction.

Recall that, by Definition \ref{def:PoiOperator}, we need to show that, if $\ast$ satisfies the relevant semantic properties, then there exists a non-prioritised POI revision operator $\circ$ such that $\mathbb{N}(\ast, \circ)$.

The construction works as follows: From each $\ast$ we can construct $\circ$ by setting, for all $x,y \in W$:
\[
x \preceq_{\Psi \circ A} y\
\textrm{iff }
x \preceq_{\Psi \ast A \vee \neg(x\vee y)} y
\]
Note that given \EquivalenceRevR, \CRevR{1} and  \CRevR{2}  this is equivalent to:
\[
x \preceq_{\Psi ^\circ A} y\
\textrm{iff }
\left\{
\begin{array}{ll}
x \preceq_{\Psi} y & \textrm{if } x \sim^A y \\
x \preceq_{\Psi \ast \neg y} y & \textrm{if } x \triangleleft^A y \\
x \preceq_{\Psi \ast \neg x} y & \textrm{if } y \triangleleft^A x\end{array}
\right.
\]

\noindent where (i) $x\trianglelefteq^A y$ iff $x\in\mods{A}$ or $y\in\mods{\neg A}$, (ii) $x\sim^A y$ when $x\trianglelefteq^A y$ and $y\trianglelefteq^A x$, and (iii) $x\triangleleft^A y$ when $x\trianglelefteq^A y$ but not $y\trianglelefteq^A x$.

We will establish the result by proving two main lemmas: first, we will show that $\circ$ is a non-prioritised POI revision operator (Lemma \ref{lem:CircisPOI}) and then we will show that $\mathbb{N}(\ast, \circ)$ (Lemma \ref{lem:AstIsNCirc}).

\begin{lem}\label{lem:CircisPOI}
$\circ$ is a non-prioritised POI revision operator
\end{lem}

\noindent We show that $\circ$ satisfies each of  \CCircR{1}, \CCircR{2}, \IndCircR,  \BetaCircPlusR{1} and \BetaCircPlusR{2}, as well as the requirement that $\preceq_{\Psi\circ A}$ is a TPO over $W$.
%
%
%
%
%
%
%
%
Before doing so, however, we first establish the following useful auxiliary lemma:

\begin{lem}
Let $x, y, z$ be distinct worlds such that $y\preceq_{\Psi} z$. Then the following are equivalent:
\begin{itemize} \label{lem:EquivforTrans}
\item[(i)] If $x\preceq_{\Psi \ast \neg y} y$, then $x\preceq_{\Psi \ast \neg z} z$

\item[(ii)] If $x\preceq_{\Psi \ast x \vee z} y$, then $x\preceq_{\Psi \ast x \vee y} z$

\end{itemize}
\end{lem}
\noindent The proof of Lemma \ref{lem:EquivforTrans} is as follows:
\begin{itemize}
\item[{\bf (a)}] {\bf From (i) to (ii):} Suppose $z\prec_{\Psi \ast x \vee z} x$. Then by \GammaRevR{2}  and \EquivalenceRevR, we have $z\prec_{\Psi \ast \neg z} x$. Hence $y\prec_{\Psi \ast \neg y} x$, by (i). From $z\prec_{\Psi \ast \neg z} x$ we also know that $z\prec_{\Psi} x$ by \CRevR{3}. Hence $x\notin\min(\preceq, \mods{x\vee z})$ and so, from $y\prec_{\Psi \ast \neg y} x$, we can conclude $y\prec_{\Psi \ast x \vee z} x$ by \BetaRevR{2}.

\item[{\bf (b)}] {\bf From (ii) to (i):} Suppose $z\preceq_{\Psi \ast \neg z}x$. Then $z\preceq_{\Psi}x$ by \CRevR{3}, so, since we can also assume $y\preceq_{\Psi}z$, $x\preceq_{\Psi}z$ and therefore $x\notin \min(\preceq, \mods{x\vee y})$. Then, from this and $z\prec_{\Psi\ast \neg z} x$, we obtain $z\prec_{\Psi \ast x \vee y} x$ by postulate \BetaRevR{2}. Hence from (ii), $y\prec_{\Psi \ast x \vee z} x$ and then $y\prec_{\Psi \ast \neg y} x$ by \GammaRevR{2}  and \EquivalenceRevR. \end{itemize}

\noindent We now return to the proof of Lemma \ref{lem:CircisPOI}. 
\begin{itemize}

\item[{\bf (a)}] {\bf Regarding $\preceq_{\Psi\circ A}$'s being a TPO over $W$:} We have $x\preceq_{\Psi \circ A} y \Leftrightarrow x\preceq_{\Psi \ast A \vee \neg (x \vee y)} y$. So completeness of $\preceq_{\Psi \circ A}$ follows from completeness of $\preceq_{\Psi \ast A \vee \neg (x \vee y)}$ and \EquivalenceRevR. To show that $\preceq_{\Psi \circ A}$ is transitive (i.e.~that, if $x\preceq_{\Psi \circ A} y$ and  $y\preceq_{\Psi \circ A} z$, then $x\preceq_{\Psi \circ A} z$), we go through the 8 cases according to whether each of $x, y,$ and $ z$ is in $\mods{A}$ or not:
\begin{itemize}

\item[] \begin{itemize}

\item[{\bf (i)}] {\bf $x, y, z\in\mods{A}$ or $x, y, z\in\mods{\neg A}$:} Follows from transitivity for $\ast$.

\item[{\bf (ii)}] {\bf $x, y\in\mods{A}$, $z\in\mods{\neg A}$:} Then we must show that, if $x\preceq_{\Psi} y$ and  $y\preceq_{\Psi \ast \neg z} z$, then $x\preceq_{\Psi \ast \neg z} z$. Since  $x, y\in\mods{A}$ and  $z\in\mods{\neg A}$, we know that $x \neq z$ and  $y \neq z$. Then from $x\preceq_{\Psi} y$ and \CRevR{1}, we obtain $x\preceq_{\Psi \ast \neg z} y$. From the latter and $y\preceq_{\Psi \ast \neg z} z$, we then obtain  $x\preceq_{\Psi \ast \neg z} z$ by transitivity for $\ast$.

\item[{\bf (iii)}] {$x\in\mods{A}$, $y\in\mods{\neg A}$, $z\in\mods{A}$:} Then we must show that,  if $x\preceq_{\Psi \ast \neg y} y$ and  $y\preceq_{\Psi \ast \neg y} z$, then $x\preceq_{\Psi} z$. By transitivity for $\ast$, it follows, from $x\preceq_{\Psi \ast \neg y} y$ and  $y\preceq_{\Psi \ast \neg y} z$, that $x\preceq_{\Psi \ast \neg y} z$. From $x, z\in\mods{A}$ and $y\in\mods{\neg A}$, we know $x\neq y$ and $z\neq y$. So from $x\preceq_{\Psi \ast \neg y} z$ and \CRevR{1}, we obtain $x\preceq_{\Psi} z$.

\item[{\bf (iv)}] {$x\in\mods{A}$, $y, z\in\mods{\neg A}$:} Then we must show that,  if $x\preceq_{\Psi \ast \neg y} y$ and  $y\preceq_{\Psi} z$, then $x\preceq_{\Psi  \ast \neg z} z$. If $z=y$, then $x\preceq_{\Psi  \ast \neg z} z$ follows immediately from $x\preceq_{\Psi \ast \neg y} y$. So we may assume $z\neq y$. By Lemma \ref{lem:EquivforTrans}, what we must establish is then equivalent to: if $x\preceq_{\Psi \ast x \vee z} y$ and  $y\preceq_{\Psi} z$, then $x\preceq_{\Psi  \ast x \vee y} z$. Or contraposing:  if $z\prec_{\Psi  \ast x \vee y} x$ and  $y\preceq_{\Psi} z$, then $y\prec_{\Psi \ast x \vee z} x$. So assume $z\prec_{\Psi  \ast x \vee y} x$ and  $y\preceq_{\Psi} z$. Now, if $x\preceq_{\Psi} z$, then $x\preceq_{\Psi  \ast x \vee y} z$ by \CRevR{3}. So assume $z\prec_{\Psi} x$. We therefore have: $x\in\mods{x\vee y}$, $z\notin \mods{x\vee y}$, $z\prec_{\Psi  \ast x \vee y} x$, $y\preceq_{\Psi} z$ and $x\notin\min(\preceq,\mods{x\vee z})$. From this, by \AlphaRevR{2}, we can then infer that $y\prec_{\Psi \ast x \vee z} x$, as required.

\item[{\bf (v)}] {$x\in\mods{\neg A}$, $y, z\in\mods{A}$:} Then we must show that, if $x\preceq_{\Psi \ast \neg x} y$ and  $y\preceq_{\Psi} z$, then $x\preceq_{\Psi  \ast \neg x} z$. Since  $x\in\mods{\neg A}$ and $y, z\in\mods{A}$, we know that $x\neq y$ and $x\neq z$. Hence from $y\preceq_{\Psi}z$, we know $y\preceq_{\Psi  \ast \neg x} z$. The desired implication then follows from transitivity for $\ast$.

\item[{\bf (vi)}] {$x\in\mods{\neg A}$, $y\in\mods{A}$, $z\in\mods{\neg A}$:} Then we must show that, if $x\preceq_{\Psi \ast \neg x} y$ and  $y\preceq_{\Psi \ast \neg z} z$, then $x\preceq_{\Psi} z$, or, equivalently, that, if $x\preceq_{\Psi \ast \neg x} y$ and $z\prec_{\Psi} x$, then  $z\prec_{\Psi \ast \neg z} y$. So suppose $x\preceq_{\Psi \ast \neg x} y$ and $z\prec_{\Psi} x$. If $y\preceq_{\Psi} x$, then by \IndRevR~we would have $y\prec_{\Psi\ast \neg x} x$: contradiction. Hence we may assume $x\prec_{\Psi} y$. From this and $z\prec_{\Psi} x$ we have, by transitivity,  $z\prec_{\Psi} y$ and hence $y\notin\min(\preceq, \mods{y\vee z})$. From this and $x\preceq_{\Psi \ast \neg x} y$, using postulate \BetaRevR{1}, we can deduce $x\preceq_{\Psi \ast y\vee z} y$. We therefore have:  $y\in\mods{y\vee z}$, $x\notin \mods{y\vee z}$, $x\preceq_{\Psi \ast y\vee z} y$, $z\prec_{\Psi} x$ and $y\notin\min(\preceq_{\Psi},\mods{x\vee y})$. From this, by \AlphaRevR{3}, we can then infer that $z\prec_{\Psi \ast x \vee y} x$, and so $z\prec_{\Psi\ast\neg z}y$, by \GammaRevR{2}, as required.

\item[{\bf (vii)}] {$x, y\in\mods{\neg A}$, $z\in\mods{A}$:} Then we must show that, if $x\preceq_{\Psi} y$ and  $y\preceq_{\Psi \ast \neg y} z$, then $x\preceq_{\Psi \ast \neg x} z$. If $x=y$, then this holds immediately, so we may assume $x\neq y$. Now suppose $x\preceq_{\Psi} y$ and  $y\preceq_{\Psi \ast \neg y} z$. If $z\preceq_{\Psi} y$, then $z\prec_{\Psi \ast \neg y} y$ by \IndRevR: contradiction. So we may assume $y\prec_{\Psi} z$. From this and $x\preceq_{\Psi} y$, we know, by transitivity, that  $x\prec_{\Psi} z$, so $z\notin\min(\preceq_{\Psi},\mods{x \vee z})$. It then follows that $y\preceq_{\Psi\ast x\vee z} z$ by postulate \BetaRevR{1}. We therefore have:  $z\in\mods{x\vee z}$, $y\notin \mods{x\vee z}$, $y\preceq_{\Psi \ast x\vee z} z$, $x\prec_{\Psi} y$ and $z\notin\min(\preceq_{\Psi},\mods{y\vee z})$. From this, by \AlphaRevR{1}, we can then infer that $x\prec_{\Psi \ast y \vee z} z$, and so $x\preceq_{\Psi\ast\neg x}z$, by \GammaRevR{1},   as required.

\end{itemize}
\end{itemize}

\item[{\bf (b)}] {\bf Regarding \CCircR{1} \& \CCircR{2}:} we have already noted towards the beginning of the proof that $x\preceq_{\Psi \circ A} y$ iff $x\preceq_{\Psi} y$, whenever $x\sim^{A} y$.

\item[{\bf (c)}] {\bf Regarding \IndCircR:} Suppose $x\triangleleft^{A} y$ and $x\preceq_{\Psi} y$. We must show that $x\preceq_{\Psi \circ A} y$ and $y\npreceq_{\Psi \circ A} x$. For this, it suffices to show that $x\preceq_{\Psi \ast \neg y} y$ and $y\npreceq_{\Psi \ast \neg y} x$, i.e. that $x\prec_{\Psi \ast \neg y} y$. This follows from \IndRevR.

\item[{\bf (d)}] {\bf Regarding  \BetaCircPlusR{1} \& \BetaCircPlusR{2}:} Proposition 3 of \cite{booth2011revise}  tells us that, if $\circ$ satisfies  the previous properties, then  \BetaCircPlusR{1}~and \BetaCircPlusR{2}~are jointly equivalent to the following condition:
\begin{tabbing}
\=BLAHII: \=\kill
\> \IIA \> If $A$ and $B$ agree on $x$ and $y$, then  $x\preceq_{\Psi \circ A} y$ iff  $x\preceq_{\Psi \circ B} y$\\[-0.25em]
\end{tabbing} 
\vspace{-1em}
\noindent where, given $A, B\in L$ and $x,y\in W$, $A$ and $B$ are said to {\em agree} on $x$ and $y$ iff either (i) $x\triangleleft^{A} y$ and $x\triangleleft^{B} y$, (ii)  $x\sim^{A} y$ and $x\sim^{B} y$ or (iii) $y\triangleleft^{A} x$ and $y\triangleleft^{B} x$. Hence it suffices to show that $\circ$ satisfies \IIA. But this is immediate from our characterisation of $\circ$ towards the beginning of this proof:    
\begin{tabbing}
\=BLAHII: \=\kill
\> $ x\preceq_{\Psi \circ A} y = \begin{cases} x\preceq_{\Psi} y, & \mbox{if } x\sim^{A}y \\ 
x\preceq_{\Psi \ast \neg y} y, & \mbox{if } x\triangleleft^{A}y \\ 
x\preceq_{\Psi \ast \neg x} y, & \mbox{if } y\triangleleft^{A}x \end{cases} $ \\[-0.25em]
\end{tabbing} 
\vspace{-1em}

\end{itemize}

\noindent We now prove our second main lemma:
\begin{lem}\label{lem:AstIsNCirc}
$\mathbb{N}(\ast, \circ)$
\end{lem}

\noindent We require:

\begin{itemize}

\item[]  $x \preceq_{\Psi \ast A} y$ iff
\begin{itemize}

\item[(i)] $x \in \min(\preceq_{\Psi\circ A}, \mods{A})$, or 

\item[(ii)] $x, y \notin \min(\preceq_{\Psi\circ A}, \mods{A})$ and $x\preceq_{\Psi \ast A \vee \neg(x \vee y)} y$

\end{itemize}
\end{itemize}
We can however replace this with
\begin{itemize}

\item[]  $x \preceq_{\Psi \ast A} y$ iff
\begin{itemize}

\item[(i)] $x \in \min(\preceq, \mods{A})$, or 

\item[(ii)] $x, y \notin \min(\preceq, \mods{A})$ and $x\preceq_{\Psi \ast A \vee \neg(x \vee y)} y$

\end{itemize}
\end{itemize}
since $\circ$ satisfies \CCircR{1}.
\begin{itemize}

\item[{\bf (a)}] {\bf Regarding the left-to-right direction:} Suppose that $x \preceq_{\Psi \ast A} y$ and $x \notin \min(\preceq, \mods{A})$. If $y \in \min(\preceq, \mods{A})$, then $y \prec_{\Psi \ast A} x$, by Success: contradiction. Hence $y \notin \min(\preceq, \mods{A})$. It remains to be shown that $x\preceq_{\Psi \ast A \vee \neg(x \vee y)} y$. If $x\sim^{A}y$, then the conclusion follows by \CRevR{1}--\CRevR{2}. If $y\triangleleft^{A}x$, then the conclusion follows from $x \preceq_{\Psi \ast A} y$ and \GammaRevR{1}. Finally, if $x\triangleleft^{A}y$, then $x\in\mods{A\vee \neg(x \vee y)}$ and  $y\in\mods{\neg (A\vee \neg(x \vee y))}$. Together with $x\notin\min(\preceq_{\Psi},\mods{A})$ and $x \preceq_{\Psi \ast A} y$, the desired conclusion then follows by postulate \BetaRevR{2}.

\item[{\bf (b)}] {\bf Regarding the right-to-left direction:} If $x \in \min(\preceq, \mods{A})$, then  $x \preceq_{\Psi \ast A} y$ by Success. So suppose $x, y \notin \min(\preceq, \mods{A})$ and $x\preceq_{\Psi \ast A \vee \neg(x \vee y)} y$. We must show $x \preceq_{\Psi \ast A} y$. If $x\sim^{A}y$, then the conclusion follows by \CRevR{1}--\CRevR{2}. If $x\triangleleft^{A}y$, then the conclusion follows by \GammaRevR{2}. Finally, if $y\triangleleft^{A}x$, then the conclusion follows from postulate \BetaRevR{1}. \qed

\end{itemize}

\end{newproof}

\vspace{1em}


\SyntacticOmega*

\begin{newproof} The equivalence of \CRevS{1} and \CRevS{2} to  \CRevR{1} and \CRevR{2} is well known. So we first show that, given  \IndRevR, \AlphaRevR{i}~entails  \OmegaRevS{i}, for $1\leq i \leq 3$.
\begin{itemize}

\item[{\bf (i)}] {\bf Regarding \OmegaRevS{1}:} From $A \not\in \bel{(\Psi \ast A) \ast B}$ we know there exists $y \in \mods{\neg A} \cap \min(\preceq_{\Psi \ast A}, \mods{B})$. From $\neg A \not\in \bel{\Psi \ast A \vee B}$ there exists $z \in \mods{A} \cap \min(\preceq_\Psi, \mods{A \vee B})$. From the minimality of $z$ we know $z\preceq_\Psi y$. If it were the case that $z \in \mods{B}$ then $y \preceq_{\Psi \ast A} z$ by the minimality of $y$ and so we must have $y \prec_\Psi z$ by \IndRevR--contradicting the minimality of $z$. Hence $z \in \mods{\neg B}$. Now assume for contradiction $B \in \bel{(\Psi \ast B) \ast A}$ and let $x \in \min(\preceq_{\Psi \ast B}, \mods{A})$. Then $x \in \mods{B}$ and, since $z \in \mods{A \wedge \neg B}$, $x \prec_{\Psi \ast B} z$. Since $y \in \min(\preceq_{\Psi \ast A}, \mods{B})$ we have $y \preceq_{\Psi \ast A} x$ and so also $y \prec_\Psi x$ by \IndRevR~which gives $x \not\in \min(\preceq_\Psi, \mods{B})$. We have now established $x \in \mods{A}$, $y \in \mods{\neg A}$, $y \preceq_{\Psi \ast A} x$, $z \preceq_\Psi y$ and $x \not\in \min(\preceq_\Psi, \mods{B})$. Hence we may deduce, by \AlphaRevR{1}, that $z \preceq_{\Psi \ast B} x$, contradicting what we already established. Hence $B \not\in \bel{(\Psi \ast B) \ast A}$.

\item[{\bf (ii)}] {\bf Regarding \OmegaRevS{2}:} Assume for contradiction $\neg B \not\in \bel{(\Psi \ast B) \ast A}$. Then there exists $x \in \mods{B} \cap \min(\preceq_{\Psi \ast B}, \mods{A})$. From $\neg A \in \bel{(\Psi \ast A) \ast B}$ we know $x \not\in \min(\preceq_{\Psi \ast A}, \mods{B})$. Let $y \in \min(\preceq_{\Psi \ast A}, \mods{B})$. Then $y \prec_{\Psi \ast A} x$ and $y \in \mods{\neg A}$. From $y \prec_{\Psi \ast A} x$ we also know $y \prec_{\Psi} x$ by \CRevR{4} (which follows from \AlphaRevR{2}), so $x \not\in \min(\preceq_\Psi, \mods{B})$. From $\neg A \not\in \bel{\Psi \ast A \vee B}$ there exists $z \in \mods{A} \cap \min(\preceq_\Psi, \mods{A \vee B})$. Since $y \in \mods{B}$ we have $z \preceq_\Psi y$. 
So we have established $x \in \mods{A}$, $y \in \mods{\neg A}$, $y \prec_{\Psi \ast A} x$, $z \preceq_\Psi y$ and $x \not\in \min(\preceq_\Psi, \mods{B})$. We can then apply \AlphaRevR{2}~to deduce $z \prec_{\Psi \ast B} x$, contradicting the minimality of $x$. Hence $\neg B \in \bel{(\Psi \ast B) \ast A}$ as required.

\item[{\bf (iii)}] {\bf Regarding \OmegaRevS{3}:} From $A \not\in \bel{(\Psi \ast A) \ast B}$ there exists $y \in \mods{\neg A} \cap \min(\preceq_{\Psi \ast A}, \mods{B})$. Assume for contradiction $\neg B \not\in \bel{(\Psi \ast B) \ast A}$. Then there exists $x \in \mods{B} \cap \min(\preceq_{\Psi \ast B}, \mods{A})$. By the minimality of $y$ we know $y \preceq_{\Psi \ast A} x$. Since $y \in \mods{\neg A}$ and $x \in \mods{A}$ this in turn gives $y \prec_\Psi x$ by \IndRevR, so $x \not\in \min(\preceq_\Psi, \mods{B})$. Since $y \in \mods{B}$ and $\neg B \in \bel{\Psi \ast A \vee B}$ there must exist some $z \in \mods{A \wedge \neg B}$ such that $z \prec_\Psi y$.
So we have established $x \in \mods{A}$, $y \in \mods{\neg A}$, $y \preceq_{\Psi \ast A} x$, $z \prec_\Psi y$ and $x \not\in \min(\preceq_\Psi, \mods{B})$. Hence we may apply \AlphaRevR{3}~and deduce $z \prec_{\Psi \ast B} x$, contradicting the minimality of $x$. Hence $\neg B \in \bel{(\Psi \ast B) \ast A}$ as required. 

\end{itemize}
\noindent Assuming AGM in the background, we now first show that, given \EquivalenceRevS, \CRevS{1}, \CRevS{2}, \BetaRevS{1} and \BetaRevS{2}, \OmegaRevS{i} entails \AlphaRevR{i}, for $1\leq i \leq 3$. We then show that \OmegaRevS{1} entails \IndRevR. 
\begin{itemize}

\item[{\bf (a)}]  
\begin{itemize}

\item[{\bf (i)}] {\bf Regarding \AlphaRevR{1}:} First note that from the assumptions we already obtain $y \preceq_{\Psi \ast C} x$ from \BetaRevR{1}. If $y = z$ then this clearly gives us the required conclusion, so we may assume $y \neq z$.
Now, from $x \in \mods{A}$, $y \in \mods{\neg A}$ and $y \preceq_{\Psi \ast A} x$, we know $y \prec_\Psi x$, by \IndRevS. Hence we have established $z \preceq_\Psi y \prec_\Psi x$. If $z \in \mods{C}$ or $x \in \mods{\neg C}$, then from $z \prec_\Psi x$ we obtain $z \prec_{\Psi \ast C} x$ from \CRevR{1}, \CRevR{2} or \CRevR{3} (which follows from \IndRevR) and so we obtain the required conclusion $z \preceq_{\Psi \ast C} x$. So assume $z \in \mods{\neg C}$ and $x \in \mods{C}$. If $y \in \mods{\neg C}$, then, from $z \in \mods{\neg C}$ and $z \preceq_\Psi y$, we obtain $z \preceq_{\Psi \ast C} y$ by \CRevR{2}, so the required conclusion follows from this, given $y \preceq_{\Psi \ast C} x$ and transitivity. So assume $y \in \mods{C}$. If $z \in \mods{\neg A}$, then, since $y \in \mods{\neg A}$, we obtain $z \preceq_{\Psi \ast A} y$, by \CRevR{2}. So $z \preceq_{\Psi \ast A} x$ by transitivity with the assumption $y \preceq_{\Psi \ast A} x$. We can then apply \BetaRevR{1}, using this together with the assumptions $x \in \mods{A}$, $z \in \mods{\neg A}$ and $x \not\in \min(\preceq_\Psi, \mods{C})$, to obtain the desired result that $z \preceq_{\Psi \ast C} x$. So assume $z \in \mods{A}$. We have now built up the following assumptions about $x,y,z$: (i) $x \in \mods{A \wedge C}$, (ii) $y \in \mods{\neg A \wedge C}$, and (iii) $z \in \mods{A \wedge \neg C}$. To show the desired result that $z \preceq_{\Psi \ast C} x$ in this final case, it suffices, by \GammaRevR{1} and \EquivalenceRevR, to show $z \preceq_{\Psi \ast x \vee y} x$, which is equivalent to $x \vee y \not\in [(\Psi \ast x \vee y) \ast x \vee z]$ (since we assume $z \neq y$ and we know also $z \neq x$ from $z \prec_\Psi x$). To prove this it suffices, by \OmegaRevS{1}~and \EquivalenceRevR, to show $\neg(x \vee z) \not\in [\Psi \ast x \vee y \vee z]$ and $x \vee z \not\in [(\Psi \ast x \vee z) \ast x \vee y]$. But the former holds since we have already established $z \preceq_\Psi y \prec_\Psi x$, while the latter is equivalent to $y \preceq_{\Psi \ast x \vee z} x$. This will follow from $y \preceq_{\Psi \ast A} x$ and \BetaRevR{1}, provided we have $x \not\in \min(\preceq_\Psi, \mods{x \vee z})$, i.e.~$z \prec_\Psi x$. But we have already established that. 

\item[{\bf (ii)}] {\bf Regarding \AlphaRevR{2}:}  First note that, from the assumptions, we already obtain $y \prec_{\Psi \ast C} x$ from \BetaRevR{2}. If $y = z$, then this clearly gives us the required conclusion. So we may assume $y \neq z$.
Now, from $x \in \mods{A}$, $y \in \mods{\neg A}$ and $y \prec_{\Psi \ast A} x$, we know $y \prec_\Psi x$, by \CRevR{4}. Hence we have established $z \preceq_\Psi y \prec_\Psi x$. If $z \in \mods{C}$ or $x \in \mods{\neg C}$, then from $z \prec_\Psi x$ we obtain $z \prec_{\Psi \ast C} x$, by \CRevR{1}, \CRevR{2} or \CRevR{3}, as required. So assume $z \in \mods{\neg C}$ and $x \in \mods{C}$. If $y \in \mods{\neg C}$, then, from $z \in \mods{\neg C}$ and $z \preceq_\Psi y$, we obtain $z \preceq_{\Psi \ast C} y$, by \CRevR{2}. So the required conclusion follows from this with $y \prec_{\Psi \ast C} x$ and transitivity. So assume $y \in \mods{C}$. If $z \in \mods{\neg A}$ then, since $y \in \mods{\neg A}$, we obtain $z \preceq_{\Psi \ast A} y$ by \CRevR{2}. So $z \prec_{\Psi \ast A} x$, by transitivity, with the assumption $y \prec_{\Psi \ast A} x$. We can then apply \BetaRevR{2}, using this together with the assumptions $x \in \mods{A}$, $z \in \mods{\neg A}$ and $x \not\in \min(\preceq_\Psi, \mods{C})$, to obtain the desired $z \prec_{\Psi \ast C} x$. So assume $z \in \mods{A}$. We now have built up the following assumptions about $x,y,z$: (i) $x \in \mods{A \wedge C}$, (ii) $y \in \mods{\neg A \wedge C}$, and  (iii) $z \in \mods{A \wedge \neg C}$. To show the desired result that $z \prec_{\Psi \ast C} x$ in this final case, it suffices, by \GammaRevR{2}~and \EquivalenceRevR, to show $z \prec_{\Psi \ast x \vee y} x$, which is equivalent to $\neg(x \vee y) \in [(\Psi \ast  x \vee y) \ast x \vee z]$ (since we assume $z \neq y$ and we know also $z \neq x$ from $z \prec_\Psi x$). To prove this it suffices, by \OmegaRevS{2} and \EquivalenceRevR, to show $\neg(x \vee z) \not\in [\Psi \ast x \vee y \vee z]$ and $\neg(x \vee z) \in [(\Psi \ast x \vee z) \ast x \vee y]$. But the former holds since we already established $z \preceq_\Psi y \prec_\Psi x$, while the latter is equivalent to $y \prec_{\Psi \ast x \vee z} x$. This will follow from $y \prec_{\Psi \ast A} x$ and \BetaRevR{2}, provided we have $x \not\in \min(\preceq_\Psi, \mods{x \vee z})$, i.e.~$z \prec_\Psi x$. But we have already established  that. 

\item[{\bf (iii)}] {\bf Regarding \AlphaRevR{3}:} From $x \in \mods{A}$, $y \in \mods{\neg A}$ and $y \preceq_{\Psi \ast A} x$, we know $y \preceq_\Psi x$ by \CRevR{3}. Hence we have established $z \prec_\Psi y \preceq_\Psi x$. If $z \in \mods{C}$ or $x \in \mods{\neg C}$, then, from $z \prec_\Psi x$, we obtain $z \prec_{\Psi \ast C} x$,  by \CRevR{1}, \CRevR{2}~or \CRevR{3}, as required. So assume $z \in \mods{\neg C}$ and $x \in \mods{C}$. From the assumptions, we already know $y \preceq_{\Psi \ast C} x$, by \BetaRevR{1}. If $y \in \mods{\neg C}$, then,  from $z \in \mods{\neg C}$ and $z \prec_\Psi y$, we obtain $z \prec_{\Psi \ast C} y$, by \CRevR{2}. So the required conclusion follows from this, given $y \preceq_{\Psi \ast C} x$ and transitivity. So assume $y \in \mods{C}$. If $z \in \mods{\neg A}$, then, since $y \in \mods{\neg A}$, we obtain $z \prec_{\Psi \ast A} y$, by \CRevR{2}. So $z \prec_{\Psi \ast A} x$,  by transitivity, alongside the assumption $y \preceq_{\Psi \ast A} x$. We can then apply \BetaRevR{2}, using this together with the assumptions $x \in \mods{A}$, $z \in \mods{\neg A}$ and $x \not\in \min(\preceq_\Psi, \mods{C})$, to obtain the desired result that $z \prec_{\Psi \ast C} x$. So assume $z \in \mods{A}$. We  have now built up the following assumptions about $x,y,z$: (i) $x \in \mods{A \wedge C}$, (ii) $y \in \mods{\neg A \wedge C}$, and (iii) $z \in \mods{A \wedge \neg C}$. To show the desired result that $z \prec_{\Psi \ast C} x$ in this final case, it suffices, by \GammaRevR{2}~and \EquivalenceRevR, to show $z \prec_{\Psi \ast x \vee y} x$, which is equivalent  to $\neg(x \vee y) \in \bel{(\Psi \ast x \vee y ) \ast  x \vee z}$ (since $y \neq z \neq x$ from $z \prec_\Psi y \preceq_\Psi x$). To prove this, it suffices, by \OmegaRevS{3}~and \EquivalenceRevR, to show $\neg(x \vee y) \in \bel{\Psi \ast x \vee y \vee z}$ and $x \vee z \not\in \bel{(\Psi \ast x \vee z) \ast x \vee y}$. But the former holds, since we already established $z \prec_\Psi y \preceq_\Psi x$, while the latter is equivalent to $y \preceq_{\Psi \ast x \vee z} x$. This will follow from $y \preceq_{\Psi \ast A} x$ and \BetaRevR{1}, provided we have $x \not\in \min(\preceq_\Psi, \mods{x \vee z})$, i.e.~$z \prec_\Psi x$. But we have already established that. 
\end{itemize}

\item[{\bf (b)}] {\bf Regarding \IndRevR:} 
We will show that \OmegaRevS{1} implies \IndRevS, whose equivalence to \IndRevR~is well known. So suppose $\neg A \not\in [\Psi \ast B]$. We must show $A \in [(\Psi \ast A) \ast B]$. From $\neg A \not\in [\Psi \ast B]$ and the AGM postulates, we obtain $\neg A \not\in \bel{\Psi \ast A \vee B}$ and also $\bel{\Psi \ast B} \subseteq \bel{(\Psi \ast B) \ast A}$. Since $B \in \bel{\Psi \ast B}$, by Success, the latter gives us $B \in \bel{(\Psi \ast B) \ast A}$. Then, from this and $\neg A \not\in \bel{\Psi \ast A \vee B}$, we obtain the required $A \in [(\Psi \ast A) \ast B]$ by \OmegaRevS{1}. \qed

\end{itemize}

\end{newproof}

\vspace{1em}


\AlphaSyntactic*

\begin{newproof}

\begin{itemize} 

\item[{\bf (a)}]

\begin{itemize} 

\item[{\bf (i)}] {\bf From \BetaRevR{3} to \BetaRevS{3}:} Assume $B_2 \notin \bel{\Psi \ast B_1}$, $B_1\rightarrow A\notin \bel{(\Psi \ast A) \ast B_2}$  and $B_2\rightarrow \neg A\in \bel{\Psi \ast C}$. From the first assumption, $\exists z\in \min(\preceq_{\Psi}, \mods{B_1})\cap \mods{\neg B_2}$ and,  from the second, $\exists y\in \min(\preceq_{\Psi \ast A}, \mods{B_2})\cap \mods{B_1\wedge\neg A}$. From this, we have $y\in\mods{B_1}$ and $z\in \min(\preceq_{\Psi}, \mods{B_1})$, hence: (1) $z \preceq_{\Psi} y$. 

Assume now for reductio, the negation of the consequent of \BetaRevS{3}, so that  $\min(\preceq_{\Psi \ast C}, \mods{B_1\veebar B_2})\subseteq \mods{B_2 \wedge A}$. So we have $\exists x \in \min(\preceq_{\Psi \ast C}, \mods{B_1\veebar B_2})\cap \mods{B_2 \wedge A}$. From this, which entails $x\in\mods{B_2}$, and the facts that $y\in \min(\preceq_{\Psi \ast A}, \mods{B_2})$ and $x\in\mods{B_2}$, we recover: (2) $y \preceq_{\Psi \ast A} x$. From our third initial assumption that $B_2\rightarrow \neg A\in \bel{\Psi \ast C}$ and the fact that $x\in  \mods{B_2 \wedge A}$, we obtain: (3) $x \notin \min(\preceq_{\Psi}, \mods{C})$. Since $z\in \mods{\neg B_2}$ and $y\in \mods{B_2}$, we also know: (4) $z\neq y$.

As we already know that $x\in\mods{A}$ and $y\in\mods{\neg A}$, (1), (2), (3) and (4) enable us to apply \BetaRevR{3} to infer $z \preceq_{\Psi \ast C} x$.  Given that we know that $z \in \mods{B_1 \wedge \neg B_2}$ and $x \in \min(\preceq_{\Psi \ast C}, \mods{B_1\veebar B_2})$, we can conclude from this last proposition that $z \in \min(\preceq_{\Psi \ast C}, \mods{B_1 \veebar B_2})$. But we also know that $z \in  \mods{\neg B_2}$. So we can conclude that $\min(\preceq_{\Psi \ast C}, \mods{B_1\veebar B_2})\nsubseteq \mods{B_2\wedge A}$ after all, as required.

\item[{\bf (ii)}] {\bf From \BetaRevS{3} to \BetaRevR{3}:} Assume the antecedent of  \BetaRevR{3}: $x \notin \min(\preceq, \mods{C})$, $x \in \mods{A}$, $y \in \mods{\neg A}$, $z \preceq y$, $y \preceq_{\Psi\ast A} x$, and $z \neq y$.

If $z = x$, then it follows from this, by reflexivity of $\preceq$ that $z \preceq_{\Psi\ast C} x$ and we are done. So assume henceforth that $z \neq x$.

Assume for reductio that  $\neg (x \vee y) \vee \neg A \notin \bel{\Psi \ast C}$, so that $\exists w \in \min(\preceq_{\Psi}, \mods{C}) \cap \mods{A\wedge(x \vee y)}$. From $x \in \mods{A}$ and $y \in \mods{\neg A}$, we have $\mods{A\wedge(x \vee y)} = \{x\}$. This means that $x \in \min(\preceq, \mods{C})$. But we initially assumed this to be false. So we can conclude by reductio: (1) $\neg (x \vee y) \vee \neg A \in \bel{\Psi \ast C}$.

Since $z \preceq y$, $z \neq y$ and $z \neq x$: (2) $x \vee y  \notin \bel{\Psi \ast z \vee y}$. Furthermore, from $y \in \mods{\neg A}$ and $y \preceq_{\Psi\ast A} x$, we can infer: (3) $\neg (z \vee y) \vee A \notin \bel{(\Psi \ast A) \ast x \vee y}$.

(1), (2) and (3) then enable us to apply \BetaRevS{3}, with $B_1 = z \vee y$ and $B_2 = x \vee y$, to recover $(x \vee y) \wedge A \notin \bel{(\Psi \ast C) \ast (z \vee y) \veebar (x \vee y)}$. Given \EquivalenceRevR, this allows us to infer $(x \vee y) \wedge A \notin \bel{(\Psi \ast C) \ast z \vee x}$, from which it follows, by $x \in \mods{A}$, that $z \preceq_{\Psi\ast C} x$, as required.

\end{itemize} 

\item[{\bf (b)}]

\begin{itemize} 

\item[{\bf (i)}] {\bf From \BetaRevR{4} to \BetaRevS{4}:} Assume the antecedent of  \BetaRevS{4}: $B_2 \notin \bel{\Psi \ast B_1}$, $B_1\wedge \neg A\in \bel{(\Psi \ast A) \ast B_2}$ and  $B_2\rightarrow \neg A\in \bel{\Psi \ast C}$.  From the first assumption, $\exists z\in \min(\preceq_{\Psi}, \mods{B_1})\cap \mods{\neg B_2}$  and, from the second, $\min(\preceq_{\Psi \ast A}, \mods{B_2})\subseteq \mods{B_1\wedge\neg A}$. 

Consider now an arbitrary $y\in \min(\preceq_{\Psi \ast A}, \mods{B_2})$. By the previous inclusion, we have  $y\in\mods{B_1}$, and so, since $z\in \min(\preceq_{\Psi}, \mods{B_1})$: (1) $z \preceq_{\Psi} y$.

Assume now for reductio, the negation of the consequent of \BetaRevS{4}, so that $\min(\preceq_{\Psi \ast C}, \mods{B_1\veebar B_2})\nsubseteq \mods{B_2 \rightarrow \neg A}$. From this, $\exists x \in \min(\preceq_{\Psi \ast C}, \mods{B_1\veebar B_2})\cap \mods{B_2 \wedge A}$. It follows from this that $x\in\mods{B_2}$ and $x\in\mods{A}$ and we already know that $\min(\preceq_{\Psi \ast A}, \mods{B_2})\subseteq \mods{\neg A}$. Hence: (2) $y \prec_{\Psi \ast A} x$. From our third initial assumption and the fact that $x\in  \mods{B_2 \wedge A}$ we can also infer: (3) $x \notin \min(\preceq_{\Psi}, \mods{C})$. Furthermore,  since $z\in \mods{\neg B_2}$ and $y\in \mods{B_2}$, it follows that (4) $z\neq y$.

As we already know that $x\in\mods{A}$ and $y\in\mods{\neg A}$, (1), (2), (3) and (4) enable us to apply \BetaRevR{4} to infer $z \prec_{\Psi \ast C} x$. From this, since $z \in \mods{B_1 \wedge \neg B_2}$, it follows that  $x \notin \min(\preceq_{\Psi \ast C}, \mods{B_1 \veebar B_2})$. But this contradicts our assumption that $x \in \min(\preceq_{\Psi \ast C}, \mods{B_1\veebar B_2})$, so we can conclude, by reductio, that 
 $\min(\preceq_{\Psi \ast C}, \mods{B_1\veebar B_2})\subseteq \mods{B_2\rightarrow \neg A}$, as required. 
 
\item[{\bf (ii)}] {\bf From \BetaRevS{4} to \BetaRevR{4}:} Assume the antecedent of  \BetaRevR{4}: $x \notin \min(\preceq, \mods{C})$,  $x \in \mods{A}$, $y \in \mods{\neg A}$, $z \preceq y$, $y \prec_{\Psi\ast A} x$ and $z \neq y$ 

Assume for reductio that $z = x$. Then, by $z \preceq y$, we have $x \preceq y$. Note that, additionally, we have assumed $y \prec_{\Psi\ast A} x$. However, $x \in \mods{A}$ and $y \in \mods{\neg A}$ give us, by \CRevR{4}: If $x \preceq y$ then $x \preceq_{\Psi\ast A} y$. Contradiction. So we can conclude, by reductio, that  $z \neq x$.

Assume now for reductio that $\neg (x \vee y) \vee \neg A \notin \bel{\Psi \ast C}$, so that $\exists w \in \min(\preceq_{\Psi}, \mods{C}) \cap \mods{A\wedge(x \vee y)}$. From $x \in \mods{A}$ and $y \in \mods{\neg A}$, we already have $\mods{A\wedge(x \vee y)} = \{x\}$. So we can infer that  $x \in \min(\preceq, \mods{C})$, contradicting our initial assumption. So we can conclude, by reductio, that (1) $\neg (x \vee y) \vee \neg A \in \bel{\Psi \ast C}$, after all.

From  $z \preceq y$,  $z \neq y$ and $z \neq x$, we recover: (2) $x \vee y  \notin \bel{\Psi \ast z \vee y}$. From $y \in \mods{\neg A}$ and $y \prec_{\Psi\ast A} x$ we can infer: (3) 
$(z \vee y) \wedge \neg A \in \bel{(\Psi \ast A) \ast x \vee y}$. 

(1), (2) and (3) then enable us to apply \BetaRevS{4}, with $B_1 = z \vee y$ and $B_2 = x \vee y$, to recover $(x \vee y) \rightarrow \neg A \in \bel{(\Psi \ast C) \ast (z \vee y) \veebar (x \vee y)}$. Given \EquivalenceRevR, this allows us to infer $(x \vee y) \rightarrow \neg A \in \bel{(\Psi \ast C) \ast z \vee x}$, from which it follows, by $x \in \mods{A}$, that $z \prec_{\Psi\ast C} x$, as required.

\end{itemize} 

\item[{\bf (c)}]

\begin{itemize} 

\item[{\bf (i)}] {\bf From \AlphaRevR{3} to \AlphaRevS{3}:} Assume the antecedent of  \AlphaRevS{3}: $\neg B_2 \in \bel{\Psi \ast B_1}$, $B_1\rightarrow A\notin \bel{(\Psi \ast A) \ast B_2}$ and $B_2\rightarrow \neg A\in \bel{\Psi \ast C}$. 

From the first principle, we have $ \min(\preceq_{\Psi}, \mods{B_1})\subseteq \mods{\neg B_2}$ and, from the second,  $\exists y\in \min(\preceq_{\Psi \ast A}, \mods{B_2})\cap \mods{B_1\wedge\neg A}$. Consider now an arbitrary $z \in \min(\preceq_{\Psi}, \mods{B_1})$. Since $y\in\mods{B_1}$ and $y\in\mods{B_2}$, it then follows from $ \min(\preceq_{\Psi}, \mods{B_1})\subseteq \mods{\neg B_2}$ that (1) $z \prec_{\Psi} y$. 

Assume now for reductio, the negation of the consequent of \AlphaRevS{3}, so that $\min(\preceq_{\Psi \ast C}, \mods{B_1\veebar B_2})\nsubseteq \mods{B_2 \rightarrow \neg A}$. From this, $\exists x \in \min(\preceq_{\Psi \ast C}, \mods{B_1\veebar B_2})\cap \mods{B_2 \wedge A}$.

From $y\in \min(\preceq_{\Psi \ast A}, \mods{B_2})$ and $x\in\mods{B_2}$, we can infer: (2)  $y \preceq_{\Psi \ast A} x$. From our third initial assumption that $B_2\rightarrow \neg A\in \bel{\Psi \ast C}$, since $x\in  \mods{B_2 \wedge A}$, we can derive: (3) $x \notin \min(\preceq_{\Psi}, \mods{C})$.

As we already know that $x\in\mods{A}$ and $y\in\mods{\neg A}$, (1), (2), and (3) enable us to apply \AlphaRevR{3} to infer $z \prec_{\Psi \ast C} x$. From this, since $z \in \mods{B_1 \wedge \neg B_2}$, it follows that  $x \notin \min(\preceq_{\Psi \ast C}, \mods{B_1 \veebar B_2})$. But this contradicts our assumption that $x \in \min(\preceq_{\Psi \ast C}, \mods{B_1\veebar B_2})$, so we can conclude, by reductio, that $\min(\preceq_{\Psi \ast C}, \mods{B_1\veebar B_2})\subseteq \mods{B_2\rightarrow \neg A}$, as required. 

\item[{\bf (ii)}] {\bf From \AlphaRevS{3} to \AlphaRevR{3}:} Assume the antecedent of  \AlphaRevR{4}:  (a) $x \notin \min(\preceq, \mods{C})$, (b) $x \in \mods{A}$, (c) $y \in \mods{\neg A}$, (d) $z \prec y$, and (e) $y \preceq_{\Psi\ast A} x$.

Assume for reductio that $z = x$. Then, by $z \prec y$, we have $x \prec y$. Note that additionally, we have assumed $y \preceq_{\Psi\ast A} x$. However, $x \in \mods{A}$ and $y \in \mods{\neg A}$ give us, by \CRevR{3}: If $x \prec y$ then $x \prec_{\Psi\ast A} y$. Contradiction. So we can conclude, by reductio, that  $z \neq x$.

Assume now for reductio that $\neg (x \vee y) \vee \neg A \notin \bel{\Psi \ast C}$, so that $\exists w \in \min(\preceq_{\Psi}, \mods{C}) \cap \mods{A\wedge(x \vee y)}$. From $x \in \mods{A}$ and $y \in \mods{\neg A}$, we already have $\mods{A\wedge(x \vee y)} = \{x\}$. So we can infer that  $x \in \min(\preceq, \mods{C})$, contradicting our initial assumption. So we can conclude, by reductio, that (1) $\neg (x \vee y) \vee \neg A \in \bel{\Psi \ast C}$, after all.

From $z \prec y$ and $z \neq x$, we recover: (2) $\neg (x \vee y)  \in \bel{\Psi \ast z \vee y}$. From $y \in \mods{\neg A}$ and $y \preceq_{\Psi\ast A} x$, we can also infer: (3) $(z \vee y) \rightarrow A \notin \bel{(\Psi \ast A) \ast x \vee y}$. 

(1), (2) and (3) then enable us to apply \AlphaRevS{3}, with $B_1 = z \vee y$ and $B_2 = x \vee y$, to recover $(x \vee y) \rightarrow \neg A \in \bel{(\Psi \ast C) \ast (z \vee y) \veebar (x \vee y)}$. Given \EquivalenceRevR, this allows us to infer $(x \vee y) \rightarrow \neg A \in \bel{(\Psi \ast C) \ast z \vee x}$, from which it follows, by $x \in \mods{A}$, that $z \prec_{\Psi\ast C} x$, as required. \qed

\end{itemize} 
\end{itemize} 
\end{newproof}

\vspace{1em}


\WPUSPUSyntactic*

\begin{newproof}
We establish the result by deriving the following lemma:

\begin{restatable}{lem}{WPUSPUDerivation}
\label{WPUSPUDerivation}
{\bf (a)} In the presence of \CRevR{1}  and \CRevR{2},   \GammaRevR{1} and \GammaRevR{4} jointly entail:
\begin{tabbing}
\=BLAHBBB: \=\kill
\> \WPURevPlus \> If $y \preceq_{\Psi\ast A} x$ and $z \preceq_{\Psi\ast C} x$, then either    $y \preceq_{\Psi\ast A\vee C } x$ or $z \preceq_{\Psi\ast A\vee C } x$\\[-0.25em]
\end{tabbing} 
\vspace{-1em}
{\bf (b)} In the presence of \CRevR{1} and \CRevR{2}, \GammaRevR{2} and \GammaRevR{3} jointly entail: 
\begin{tabbing}
\=BLAHBBB: \=\kill
\> \SPURevPlus \> If $y \prec_{\Psi\ast A} x$ and $z \prec_{\Psi\ast C} x$, then either  $y \prec_{\Psi\ast A\vee C } x$ or $z \prec_{\Psi\ast A\vee C } x$\\[-0.25em]
\end{tabbing} 
\vspace{-1em}
\end{restatable}

\noindent Given this, the required conclusion follows immediately from Proposition 3 of \cite{BoothC16}.

\begin{itemize}

\item[{\bf (a)}] We first note that \CRevR{1}  and \GammaRevR{1} jointly entail:

\begin{tabbing}
\=BLAHII: \=\kill
\> \GammaRevPlusR{1} \>  If $x \in \mods{A}$ and $y \preceq_{\Psi \ast A} x$, then  $y \preceq_{\Psi \ast A \vee C} x$ \\[-0.25em]
\end{tabbing} 
\vspace{-1em}

From this, it follows that \WPURevPlus~holds whenever either $x \in \mods{A}$ or $x \in \mods{C}$. So assume henceforth that $x \in \mods{\neg(A \vee C)}$. 

Now assume $y \preceq_{\Psi \ast A} x$, $z \preceq_{\Psi \ast C} x$, and, for contradiction, that both $x \prec_{\Psi \ast A \vee C} y$ and $x \prec_{\Psi \ast A \vee C} z$. If either (i) $y \notin \min(\preceq, \mods{A})$ and $y \in \mods{A}$, or (ii) $y \notin \min(\preceq, \mods{A})$ and $y \in \mods{C}$
then, by \GammaRevR{4}, it follows from $x \in \mods{\neg (A \vee C)}$ and $x \prec_{\Psi \ast A \vee C} y$ that $x \prec_{\Psi \ast A} y$, contradicting our assumption that $y \preceq_{\Psi \ast A} x$. So assume that either $y \in \min(\preceq, \mods{A})$ or $y \in \mods{\neg( A \vee C)}$. By parallel reasoning from $x \in \mods{\neg (A \vee C)}$ and $x \prec_{\Psi \ast A \vee C} z$, we end up with the assumption that either 
 $z \in \min(\preceq, \mods{C})$ or $z \in \mods{\neg( A \vee C)}$.

Assume that $y \in \mods{\neg( A \vee C)}$. By \CRevR{2}, it then follows from this, $x \in \mods{\neg (A \vee C)}$ and $x \prec_{\Psi \ast A \vee C} y$ that $x \prec_{\Psi} y$. But from $x \prec_{\Psi} y$, $x, y \in \mods{\neg A}$ and \CRevR{2}  again, we have $x \prec_{\Psi \ast A} y$, contradicting our assumption that $y \preceq_{\Psi \ast A} x$.
Similarly, assuming that $z \in \mods{\neg( A \vee C)}$ leaves us with  $x \prec_{\Psi \ast C} z$, this time contradicting our assumption that $z \preceq_{\Psi \ast C} x$.

So assume that $y \notin \mods{\neg( A \vee C)}$ and $z \notin \mods{\neg( A \vee C)}$. It follows that both $y \in \min(\preceq, \mods{A})$ and $z \in \min(\preceq, \mods{C})$.

It follows from $x \prec_{\Psi \ast A \vee C} y$ that $y \notin  \min(\preceq, \mods{A\vee C})$. 
From this and $y \in \min(\preceq, \mods{A})$, we obtain $\min(\preceq, \mods{A\vee C}) = \min(\preceq, \mods{C})$. Since it also follows from $x \prec_{\Psi \ast A \vee C} z$ that $z \notin  \min(\preceq, \mods{A\vee C})$, we have $z \notin \min(\preceq, \mods{C})$, contradicting our assumption that $z \in \min(\preceq, \mods{C})$. Hence either  $y\preceq_{\Psi\ast A\vee C } x$ or $z\preceq_{\Psi\ast A\vee C } x$, as required.

\item[{\bf (b)}] We first note that \CRevR{1}  and \GammaRevR{2}  jointly entail:

\begin{tabbing}
\=BLAHII: \=\kill
\> \GammaRevPlusR{2} \>  If $x \in \mods{A}$ and $y \prec_{\Psi \ast A} x$,  $y \prec_{\Psi \ast A \vee C} x$ \\[-0.25em]
\end{tabbing} 
\vspace{-1em}

\noindent From this, it follows that \SPURevPlus~holds whenever either $x \in \mods{A}$ or $x \in \mods{C}$. So assume henceforth that $x \in \mods{\neg(A \vee C)}$. 

Now assume $y \prec_{\Psi \ast A} x$, $z \prec_{\Psi \ast C} x$, and, for contradiction, that both $x \preceq_{\Psi \ast A \vee C} y$ and $x \preceq_{\Psi \ast A \vee C} z$. If either (i) $y \notin \min(\preceq, \mods{A})$ and $y \in \mods{A}$, or (ii) $y \notin \min(\preceq, \mods{A})$ and $y \in \mods{C}$
then, by \GammaRevR{3}, it follows from $x \in \mods{\neg (A \vee C)}$ and $x \preceq_{\Psi \ast A \vee C} y$ that $x \preceq_{\Psi \ast A} y$, contradicting our assumption that $y \prec_{\Psi \ast A} x$. So assume that either $y \in \min(\preceq, \mods{A})$ or $y \in \mods{\neg( A \vee C)}$. By parallel reasoning from $x \in \mods{\neg (A \vee C)}$ and $x \prec_{\Psi \ast A \vee C} z$, we end up with the assumption that either 
 $z \in \min(\preceq, \mods{C})$ or $z \in \mods{\neg( A \vee C)}$.

Assume that $y \in \mods{\neg( A \vee C)}$. By \CRevR{2}, it then follows from this, $x \in \mods{\neg (A \vee C)}$ and $x \preceq_{\Psi \ast A \vee C} y$ that $x \preceq_{\Psi} y$. But from $x \preceq_{\Psi} y$, $x, y \in \mods{\neg A}$ and \CRevR{2}  again, we have $x \preceq_{\Psi \ast A} y$, contradicting our assumption that $y \prec_{\Psi \ast A} x$.
Similarly, assuming that $z \in \mods{\neg( A \vee C)}$ leaves us with  $x \preceq_{\Psi \ast C} z$, this time contradicting our assumption that $z \prec_{\Psi \ast C} x$.

So assume that $y \notin \mods{\neg( A \vee C)}$ and $z \notin \mods{\neg( A \vee C)}$. It follows that both $y \in \min(\preceq, \mods{A})$ and $z \in \min(\preceq, \mods{C})$.

From the fact that $x \in \mods{\neg(A \vee C)}$, it follows from $x \prec_{\Psi \ast A \vee C} y$ that $y \notin  \min(\preceq, \mods{A\vee C})$. 
From this and $y \in \min(\preceq, \mods{A})$, we obtain $\min(\preceq, \mods{A\vee C}) = \min(\preceq, \mods{C})$. Since it also follows, by $x \in \mods{\neg(A \vee C)}$, from $x \prec_{\Psi \ast A \vee C} z$ that $z \notin  \min(\preceq, \mods{A\vee C})$, we have $z \notin \min(\preceq, \mods{C})$, contradicting our assumption that $z \in \min(\preceq, \mods{C})$. Hence either  $y\prec_{\Psi\ast A\vee C } x$ or $z\prec_{\Psi\ast A\vee C } x$, as required. \qed

\end{itemize}
\end{newproof}

\vspace{1em}


\GammaOneTwoSyntactic*

\begin{newproof}
We first establish the following lemma:

\begin{restatable}{lem}{wSRrAlt}
\label{lem:wSRrAlt}
\GammaRevR{1} and \GammaRevR{2}  are respectively equivalent, in the presence of \CRevR{1}, to
\begin{tabbing}
\=BLAHII: \=\kill
\>\GammaRevPlusR{1} \> If $x \in \mods{A}$ and $y \preceq_{\Psi \ast A} x$ then $y \preceq_{\Psi \ast A\vee C } x$ \\[0.1cm]

\>\GammaRevPlusR{2} \> If $x \in \mods{A}$ and $y \prec_{\Psi \ast A} x$ then $y \prec_{\Psi \ast A\vee C } x$  \\[-0.25em]
\end{tabbing} 
\vspace{-1em}
\end{restatable}

\noindent \GammaRevPlusR{1} and \GammaRevPlusR{2} are  simply the respective strengthenings of \GammaRevR{1} and \GammaRevR{2}  in which we do not require $y \in \mods{\neg A}$ in the antecedent. So it is sufficient to show that \CRevR{1}  entails that, when $x, y \in \mods{A}$, the following both hold: (i) if $y \preceq_{\Psi \ast A} x$, then $y \preceq_{\Psi \ast A\vee C } x$ and (ii) if $y \prec_{\Psi \ast A} x$, then $y \prec_{\Psi \ast A\vee C } x$. This is an immediate consequence of the fact that \CRevR{1} entails:  If $x, y \in \mods{A}$, then $x \preceq_{\Psi\ast A} y$ iff $x \preceq_{\Psi\ast A\vee C } y$. Regarding the proof of this last implication: It follows from \CRevR{1} that,  if $x, y \in \mods{A\vee C}$, then $x \prec_{\Psi\ast A\vee C } y$ iff $x \prec_{\Psi} y$, and hence, since $\mods{A}\subseteq\mods{A\vee C}$, that:
\begin{itemize}

\item[(1)] If $x, y \in \mods{A}$, then $x \preceq_{\Psi\ast A\vee C } y$ iff $x \preceq_{\Psi} y$

\end{itemize}
But \CRevR{1} also directly gives us
\begin{itemize}

\item[(2)] If $x, y \in \mods{A}$, then $x \preceq_{\Psi\ast A} y$ iff $x \preceq_{\Psi} y$

\end{itemize}
From (1) and (2), we then recover the required result.

~

\noindent We now return to the proof of the theorem. In view of the above, we now simply need to prove equivalences between \iDI~and \iKRev{7} and \GammaRevPlusR{1} and \GammaRevPlusR{2}, respectively. Regarding \iKRev{7} and \GammaRevPlusR{2}, it is convenient here to use the following equivalent formulation of \iKRev{7}:
\begin{tabbing}
\=BLAHII: \=\kill
\>  \> $\bel{(\Psi\ast A)\ast B}\subseteq\mathrm{Cn}(\bel{(\Psi \ast A\vee C )\ast B}\cup\{A\})$ \\[-0.25em]
\end{tabbing} 
\vspace{-1em}

\noindent Here, then, is the derivation of the various implications:
\begin{itemize}

\item[{\bf (a)}]
\begin{itemize}

\item[{\bf (i)}] {\bf From \GammaRevPlusR{1} to \iDI:} Suppose $\neg A\notin \bel{(\Psi\ast A\vee C )\ast B}$. Then there exists $x\in\min(\preceq_{\Psi\ast A\vee C },\mods{B})\cap\mods{A}$. Let $y\in\min(\preceq_{\Psi\ast A},\mods{B})$. We must show $y\in\min(\preceq_{\Psi\ast A\vee C },\mods{B})$. Assume for contradiction that $y\notin\min(\preceq_{\Psi\ast A\vee C },\mods{B})$. Then $x\prec_{\Psi\ast A\vee C }y$. So, since $x\in\mods{A}$, by \GammaRevPlusR{1}, we have $x\prec_{\Psi\ast A} y$. But this contradicts $y\in\min(\preceq_{\Psi\ast A},\mods{B})$. Hence $y\in\min(\preceq_{\Psi\ast A\vee C },\mods{B})$, as required.

\item[{\bf (ii)}] {\bf From \iDI to \GammaRevPlusR{1}:} Suppose $x\in\mods{A}$ and $x\prec_{\Psi\ast A\vee C }y$. We must show $x\prec_{\Psi\ast A}y$. From $x\prec_{\Psi\ast A\vee C }y$ and $x\in\mods{A}$, we know $\neg A\notin \bel{(\Psi\ast A\vee C )\ast (x\vee y)}$. So, by \iDI, it follows that $\bel{(\Psi\ast A\vee C )\ast (x\vee y)}\subseteq \bel{(\Psi\ast A)\ast (x\vee y)}$. Moreover $x\prec_{\Psi\ast A\vee C }y$ gives us $\neg y\in\bel{(\Psi\ast A\vee C )\ast (x\vee y)}$, hence $\neg y\in\bel{(\Psi\ast A)\ast (x\vee y)}$ and therefore $x\prec_{\Psi\ast A}y$.

\end{itemize}

\item[{\bf (b)}]
\begin{itemize}

\item[{\bf (i)}] {\bf From \GammaRevPlusR{2} to \iKRev{7}:} Establishing \iKRev{7}  amounts to showing that, if $x\in\mods{\bel{(\Psi\ast A\vee C )\ast B}}\cap\mods{A}$, then $x\in\mods{\bel{(\Psi\ast A)\ast B}}$. So assume $x\in\mods{\bel{(\Psi\ast A\vee C )\ast B}}\cap\mods{A}$ and, for reductio, $x\notin\mods{\bel{(\Psi\ast A)\ast B}}$, i.e.~there exists $y\in\mods{B}$ such that $y\prec_{\Psi\ast A} x$. Since $x\in\mods{A}$,  it follows, by \GammaRevPlusR{2}, that $y\prec_{\Psi\ast A\vee C } x$ and hence, since $y\in\mods{B}$, that $x\notin\mods{\bel{(\Psi\ast A\vee C )\ast B}}$. Contradiction. Hence $x\in\mods{\bel{(\Psi\ast A)\ast B}}$, as required.

\item[{\bf (ii)}] {\bf From  \iKRev{7} to \GammaRevPlusR{2}:} Let $x \in \mods{A}$ and $x \preceq_{\Psi \ast A\vee C } y$. We must show $x \preceq_{\Psi \ast A} y$. From $x \preceq_{\Psi \ast A\vee C } y$, we know $x\in \min(\prec_{\Psi \ast A\vee C }, \{x,y\})$, i.e.~$x\in\mods{\bel{(\Psi\ast A\vee C )\ast (x\vee y)}}$. Since $x \in \mods{A}$, it follows by \iKRev{7}  that $x\in\mods{\bel{(\Psi\ast A)\ast (x\vee y)}}$, i.e.~$x\in \min(\prec_{\Psi \ast A}, \{x,y\})$. So  $x \preceq_{\Psi \ast A} y$, as required. \qed

\end{itemize}

\end{itemize}
\end{newproof}

\vspace{1em}


\KrEightSem*

\begin{newproof}
In view of Lemma \ref{lem:wSRrAlt}, we simply need to establish an equivalence between \iKRev{8}, on the one hand, and \IndRevPlusR~and  \GammaRevPlusR{1}, on the other. For convenience, we will work with the following equivalent formulation of \iKRev{8}:
\begin{tabbing}
\=BLAHII: \=\kill
\> \> If $\neg A\notin \bel{(\Psi\ast  A\vee C )\ast B}$, then  $\textrm{Cn}(\bel{(\Psi\ast  A\vee C )\ast B}\cup\{A\})$\\
\>\>$\subseteq \bel{(\Psi\ast A)\ast B}$  \\[-0.25em]
\end{tabbing} 
\vspace{-1em}
 
\begin{itemize}

\item[{\bf (i)}] {\bf From  \GammaRevPlusR{1}~and \IndRevPlusR~to \iKRev{8}:} Assume $\neg A\notin \bel{(\Psi\ast A\vee C )\ast B}$, i.e.~that there exists $x\in\min(\preceq_{\Psi\ast A\vee C },\mods{B})\cap\mods{A}$. Let $y\in\min(\preceq_{\Psi\ast A},\mods{B})$. We need to show $y\in\min(\preceq_{\Psi\ast A\vee C },\mods{B})\cap\mods{A}$. So assume for contradiction that the latter is false, i.e. that one of the following two claims is true: (1) $y\notin\min(\preceq_{\Psi\ast A\vee C },\mods{B})$, (2) $y\in\mods{\neg A}$. Assume (1). Since $x, y\in\mods{B}$, it follows that $x\prec_{\Psi\ast A\vee  C} y$. Since $x\in\mods{A}$, we then have, by \GammaRevPlusR{1}, $x\prec_{\Psi\ast A} y$, contradicting our assumption that $y\in\min(\preceq_{\Psi\ast A},\mods{B})$. Assume (2). Since $x\in\mods{A}$ and $x\preceq_{\Psi\ast A\vee  C} y$, it then follows by \IndRevPlusR~that $x\prec_{\Psi\ast A} y$, again contradicting our assumption that $y\in\min(\preceq_{\Psi\ast A},\mods{B})$. 

\item[{\bf (ii)}]  {\bf From \iKRev{8} to  \GammaRevPlusR{1} and \IndRevPlusR:} We already know that 
\begin{tabbing}
\=BLAHII: \=\kill
\> \iDI \> If $\neg A\notin \bel{(\Psi\ast A\vee C)\ast B}$,  then $\bel{(\Psi\ast A\vee C )\ast B}$\\
\>\>$\subseteq\bel{(\Psi\ast A)\ast B}$  \\[-0.25em]
\end{tabbing} 
\vspace{-1em}

\noindent which is weaker than \iKRev{8}, entails \GammaRevPlusR{1} (see proof of Proposition \ref{GammaOneTwoSyntactic}, part (a)(ii)). So we just need to recover \IndRevPlusR. So suppose $x\in\mods{A}$, $y\in\mods{\neg A}$ and $x\preceq_{\Psi\ast A\vee C } y$. We must  show $x\prec_{\Psi\ast A} y$.  From $x\in\mods{A}$ and $x\preceq_{\Psi\ast A\vee C } y$, it follows that $\neg A\notin\bel{(\Psi\ast A\vee C)\ast  x\vee y }$. So by \iKRev{8}, we have $\mathrm{Cn}(\bel{(\Psi\ast A\vee C)\ast x\vee y }\cup\{A\})\subseteq \bel{(\Psi\ast A)\ast x\vee y }$. Since $y\in\mods{\neg A}$, we have $\neg y\in\mathrm{Cn}(\bel{(\Psi\ast A\vee C )\ast x\vee y }\cup\{A\})$. Therefore $\neg y\in\bel{(\Psi\ast A)\ast x\vee y }$ and hence $x\prec_{\Psi\ast A} y$, as required. \qed

\end{itemize}
\end{newproof}


\IndPlusSoundness*

\begin{newproof}
Regarding lexicographic revision, the proof is trivial, since the latter satisfies If $x\in\mods{A}$, $y\in\mods{\neg A}$, then $x \prec_{\Psi\ast A} y$.

Regarding restrained revision: Assume $x\notin\min(\preceq,\mods{A})$, otherwise the conclusion follows by Success. Assume $y\prec_{\Psi} x$, other wise the conclusion follows by \IndRevR. From these assumptions, we have $y\prec_{\Psi \ast A\vee C} x$ by the characteristic principle of restrained revision. \qed

\end{newproof}

\vspace{1em}


\DFTwoLtoRSemantic*

\begin{newproof} For convenience, we rewrite  \GammaRevR{5} and  \GammaRevR{6} as: If $x,y \in \mods{\neg C}$ and $y \preceq_{\Psi \ast A\vee C} x$, then $y \preceq_{\Psi \ast A} x$, and, if $y \in \mods{\neg C}$ and $y \prec_{\Psi \ast A\vee C} x$, then $y \prec_{\Psi \ast A} x$
\begin{itemize}

\item[{\bf (a)}] 
\begin{itemize}

\item[{\bf (i)}] {\bf From  \GammaRevR{5} to the right-to-left direction of \iDF (i):} Assume $\min(\preceq_{\Psi \ast A \vee C}, \mods{B})\subseteq \mods{\neg C}$ and $y \in \min(\preceq_{\Psi \ast A \vee C}, \mods{B})$. Assume for reductio that $y \notin \min(\preceq_{\Psi \ast A}, \mods{B})$. Then there exists $x\in\mods{\neg C}$ such that $y \preceq_{\Psi \ast A \vee C} x$ but $x \prec_{\Psi \ast A} y$. Since $y \in \mods{\neg C}$, this contradicts \GammaRevR{5}. So $y \in \min(\preceq_{\Psi \ast A}, \mods{B})$, as required.

\item[{\bf (ii)}] {\bf From the right-to-left direction of \iDF (i) to  \GammaRevR{5}:} We consider \GammaRevR{5} contrapositively.  Assume $x,y \in \mods{\neg C}$ and $x \prec_{\Psi \ast A} y$. From $x,y \in \mods{\neg C}$, it follows that $\neg C\in\bel{(\Psi \ast A\vee C)\ast x\vee y}$. From  $x \prec_{\Psi \ast A} y$, we have $x \in \bel{(\Psi \ast A) \ast x \vee y}$. Since  $\neg C\in\bel{(\Psi \ast A\vee C)\ast x\vee y}$, by the right-to-left direction of \iDF (i), we then have $x \in \bel{(\Psi \ast A \vee C) \ast x \vee y}$. It then follows from this that $x \prec_{\Psi \ast A\vee C} y$, as required.
\end{itemize}

\item[{\bf (b)}] 
\begin{itemize}

\item[{\bf (i)}] {\bf From \GammaRevR{6} to the left-to-right direction of \iDF$\mathrm{(i)}$:} Assume $\min(\preceq_{\Psi \ast A \vee C}, \mods{B}) \subseteq \mods{\neg C}$ and $x \in \min(\preceq_{\Psi \ast A}, \mods{B})$. Assume for reductio that $x \notin \min(\preceq_{\Psi \ast A\vee C}, \mods{B})$. Then there exists $y\in\mods{\neg C}$ such that $y\prec_{\Psi\ast A\vee C} x$ but $x \preceq_{\Psi\ast A} y$, contradicting \GammaRevR{6}. Hence $x \in \min(\preceq_{\Psi \ast A\vee C}, \mods{B})$, as required.

\item[{\bf (ii)}] {\bf From the left-to-right direction of \iDF$\mathrm{(i)}$ to  \GammaRevR{6}:} Assume $y\in\mods{\neg C}$ and  $y\prec_{\Psi\ast A\vee C} x$. It follows from this that $\neg C\in\bel{(\Psi \ast A \vee C) \ast x \vee y)}$. By  the left-to-right direction of \iDF$\mathrm{(i)}$, we then have $\bel{(\Psi \ast A \vee C) \ast x \vee y}] \subseteq \bel{(\Psi \ast A) \ast x \vee y}$. Furthermore, $y\prec_{\Psi\ast A\vee C} x$ leaves us with $\neg x \in \bel{(\Psi \ast A \vee C) \ast x \vee y)}$ and hence, by the preceding inclusion $\neg x \in \bel{(\Psi \ast A) \ast x \vee y)}$. Hence $y\prec_{\Psi\ast A} x$, as required. \qed

\end{itemize}
\end{itemize}

\end{newproof}

\vspace{1em}


\DFTwoLtoRNotSoundPOI*

\begin{newproof} Since we know that restrained revision operators are POI revision operators, it will suffice to show that they violate the principle. So consider the countermodel below, where $\ast$ denotes a restrained revision operator. We have $\overline{A}\overline{C}\prec_{\Psi\ast A\vee C}\overline{A}C$, but $\overline{A}C\prec_{\Psi\ast C}\overline{A}\overline{C}$, contradicting both principles.

\bigskip

\begin{centering}
\begin{tikzpicture}[->,>=stealth']
\scalebox{1}{%
 \node[state] (K) 
 {
\begin{tabular}{c}
$A~\overline{C}$\\
$\overline{A}~\overline{C}$\\
$\overline{A}~C$\\
$A~C$\\
\end{tabular}
};

 \node[state,       
 node distance=4cm,     
 right of=K,        
 yshift=+0cm] (K*AVC)    
 {
\begin{tabular}{c}
$\overline{A}~C$\\
$A~\overline{C}$\\
$\overline{A}~\overline{C}$\\
$A~C$\\
\end{tabular}
 };

 \path (K) edge  node[anchor=north,above]
                   {
                   $\ast C$
                   } (K*AVC)
;

 \path (K)  	edge[loop left]    node[anchor=north,left]{$\ast A\vee C$} (K)

;
}
\end{tikzpicture}

\end{centering}

\qed

\end{newproof}

\vspace{1em}


\DFTwoLtoRSoundLex*

\begin{newproof} We prove the result in relation to \GammaRevR{5}, since the case of  \GammaRevR{6} is analogous. Assume $x,y\in \mods{\neg A}$ and $y \preceq_{\Psi \ast A\vee C} x$. We consider three cases:
\begin{itemize}

\item[{\bf (i) }] $x \in \mods{C}$: Assume for reductio that $y \in \mods{\neg C}$. Then, since $x\in \mods{A\vee C}$ and $y\in \mods{\neg(A\vee C)}$, lexicographic revision yields $x \preceq_{\Psi \ast A\vee C} y$. Contradiction. Hence  $y \in \mods{C}$. From $x, y\in \mods{A\vee C}$ and $y \preceq_{\Psi \ast A\vee C} x$, we obtain $y \preceq_{\Psi} x$ by \CRevR{1}, and from this, in conjunction with $x, y\in \mods{C}$ we recover the required result that $y \preceq_{\Psi \ast  C} x$, again by \CRevR{1}.

\item[{\bf (ii) }] $x \in \mods{\neg C}$ and $y \in \mods{C}$: Then, since $x\in \mods{\neg(A\vee C)}$ and $y\in \mods{A\vee C}$, lexicographic revision yields $y \preceq_{\Psi \ast A\vee C} x$, as required.

\item[{\bf (iii) }] $x, y \in \mods{\neg C}$: From $x, y\in \mods{\neg (A\vee C)}$ and $y \preceq_{\Psi \ast A\vee C} x$, we obtain $y \preceq_{\Psi} x$ by \CRevR{2}, and from this, in conjunction with $x, y\in \mods{\neg C}$ we recover the required result that $y \preceq_{\Psi \ast  C} x$, again by \CRevR{2}. \qed

\end{itemize}

\end{newproof}

\vspace{1em}


%

\end{document}